\documentclass{article}

\usepackage{arxiv}
\usepackage{arxiv_abbreviations}
\usepackage{float}
\usepackage{graphicx} 
\usepackage{amsmath}
\usepackage{amsfonts}
\usepackage{glossaries}
\usepackage{adjustbox}
\usepackage{xcolor}
\usepackage{xspace}
\usepackage{booktabs}
\usepackage{multicol}
\usepackage{multirow} 
\usepackage{hyperref}

\usepackage[capitalize]{cleveref}
\crefname{section}{Sec.}{Secs.}
\Crefname{section}{Section}{Sections}
\crefname{table}{Tab.}{Tabs.}
\Crefname{table}{Table}{Tables}

\newlength\myHeight 
\newlength\myWidth

\newacronym{ai}{AI}{Artificial Intelligence}
\newacronym{amax}{ActMax}{Activation Maximization}
\newacronym{ch}{CH}{Clever Hans}
\newacronym{cnn}{CNN}{Convolutional Neural Network}
\newacronym{nn}{NN}{Neural Network}
\newacronym{crc}{CRP}{Concept Relevance Propagation}
\newacronym{dl}{DL}{Deep Learning}
\newacronym{dnn}{DNN}{Deep Neural Network}
\newacronym{ga}{GA}{Gradient Ascent}
\newacronym{gan}{GAN}{Generative Adversarial Network}
\newacronym{hitl}{HITL}{Human in the Loop}
\newacronym{lrp}{LRP}{Layer-wise Relevance Propagation}
\newacronym{ml}{ML}{Machine Learning}
\newacronym{mlp}{MLP}{Multilayer Perceptron}
\newacronym{llm}{LLM}{Large Language Model}
\newacronym{rmax}{RelMax}{Relevance Maximization}
\newacronym{rnn}{RNN}{Recurrent Neural Network}
\newacronym{tcav}{TCAV}{Testing With Activation Vectors}
\newacronym{xai}{XAI}{eXplainable Artificial Intelligence}
\newacronym{vit}{ViT}{Vision Transformer}

\newacronym{lll}{LLL}{Low-Level hidden Layers}
\newacronym{mll}{MLL}{Mid-Level hidden Layers}
\newacronym{hll}{HLL}{High-Level hidden Layers}
\newacronym{fcl}{FCL}{Fully-Connected Layers}
\newacronym{mag}{MAG}{Magnitude Flag}

\newacronym{sem}{SEM}{Standard Error of Mean}

\hypersetup{
    colorlinks,
    linkcolor={red!60!black},
    citecolor={blue!70!black},
    urlcolor={blue!70!black}
}

\definecolor{myAcronymColor}{RGB}{0, 128, 128} 

\let\oldgls\gls
\renewcommand{\gls}[1]{\textcolor{black}{\oldgls{#1}}}

\title{Pruning By Explaining Revisited: Optimizing Attribution Methods to Prune CNNs and Transformers}

\author{Sayed Mohammad Vakilzadeh Hatefi$^1$ \and
Maximilian Dreyer$^{1}$ \and
Reduan Achtibat$^1$ \and
Thomas Wiegand$^{1,2,3}$ \and
Wojciech Samek$^{1,2,3,\dagger}$  \and
Sebastian Lapuschkin$^{1,\dagger}$ \and
\\
\\
$^1$ Fraunhofer Heinrich-Hertz-Institute, 10587 Berlin, Germany\\
$^2$ Technische Universität Berlin, 10587 Berlin, Germany\\
$^3$ BIFOLD – Berlin Institute for the Foundations of Learning and Data, 10587 Berlin, Germany\\
$^\dagger$ corresponding authors: \texttt{\{wojciech.samek,sebastian.lapuschkin\}@hhi.fraunhofer.de}\\
}

\date{} 					

\renewcommand{\shorttitle}{Optimizing Attribution Methods to Prune CNNs and Transformers}

\begin{document}
\maketitle

\begin{abstract}
	To solve ever more complex problems,
  Deep Neural Networks are scaled to billions of parameters,
  leading to huge computational costs. 
  An effective approach to reduce computational requirements and increase efficiency is to prune unnecessary components of these often over-parameterized networks.
  Previous work has shown that attribution methods from the field of eXplainable AI serve as effective means to extract and prune
  the least relevant network components in a few-shot fashion.
  We extend the current state by proposing to explicitly optimize hyperparameters of attribution methods for the task of pruning,
  and further include transformer-based networks in our analysis. 
  Our approach yields higher model compression rates of large transformer and convolutional architectures (VGG, ResNet, ViT) compared to previous works,
  while still attaining high performance on ImageNet classification tasks.
  Here, our experiments indicate that transformers have a higher degree of over-parameterization compared to convolutional neural networks. Code is available at \url{https://github.com/erfanhatefi/Pruning-by-eXplaining-in-PyTorch}.
\end{abstract}


\section{Introduction}
\label{sec:intro}

In recent years, \glspl{dnn} have been growing larger increasingly, demanding ever more computational resources and memory. 
To address these challenges, several efficient architectures, such as MobileNet~\cite{howard2017mobilenets} or EfficientFormer~\cite{li2022efficientformer}, have been proposed to reduce computational costs.
However, the gain in efficiency comes at the cost of performance, as inherently efficient architectures struggle to keep pace with the recent surge in high-performing but costly transformer models.

One widely adopted approach within the community of efficient Deep Learning is quantization~\cite{gholami2022survey, zhou2016dorefa},
which involves compressing model parameters or features by reducing the number of bits used for representation,
potentially followed by re-training to regain lost model performance. 
Another effective approach is model sparsification,
commonly referred to as pruning,
where irrelevant structures within the model are removed to reduce complexity and improve efficiency.
It is to note,
that quantization and pruning can be applied in tandem~\cite{kuzmin2024pruning}.
In this work, unlike previous studies that mandate post-retraining in their framework \cite{frankle2018lottery, peste2021ac}, we focus on pruning without additional training.

The key challenge in pruning (without sequential re-training) is the identification of structures within the model that can be removed without adversely affecting its performance. 
Earlier works have shown that structures may be chosen according to, \eg, parameter magnitudes~\cite{han2015learning,lee2021layer}, functional redundancy~\cite{geng2022pruning}, sensitivity~\cite{yang2022channel, molchanov2016pruning}, or importance in the decision-making process~\cite{yeom2021pruning,becking2020ecq}.
Especially the latter approach has been promising, utilizing tools from the seemingly unrelated field of \gls{xai} that originally seeks to explain the model reasoning process to human stakeholders.

\begin{figure}[t]
  \centering
    \includegraphics[width=0.99\linewidth]{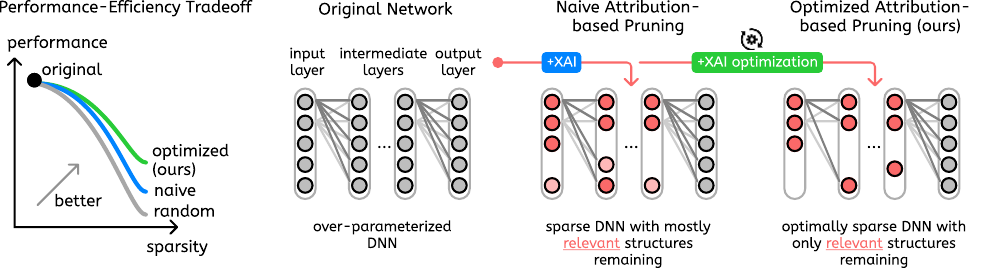}
   \caption{
   We propose a pruning framework based on optimizing attribution methods from the field of \gls{xai}.
   Compared to random pruning,
   pruning the least relevant structures first (``relevant'' according to an \gls{xai} attribution method of choice, and indicated by \emph{red color}),
   has been shown to result in an improved performance-sparsity tradeoff (simplified illustration depicted).
   By optimizing attribution methods specifically for pruning,
   we can reduce the tradeoff even further.
   }
   \label{fig:introduction} 
\end{figure}

Backpropagation-based attribution methods, particularly \gls{lrp} \cite{bach2015pixel, montavon2019layer, montavon2017explaining}, play a crucial role in this context. 
\gls{lrp} works by tracing the internal reasoning of the model, assigning contribution scores to \emph{latent} neurons, which highlight their importance to the final prediction. These scores provide a quantitative basis for identifying (ir-)relevant structures and subgraphs within the model that are (un-)essential for making accurate predictions, thus guiding the pruning process.

While prior attribution-based pruning \cite{yeom2021pruning} achieved remarkable results, it is restricted to heuristically chosen \gls{lrp} hyperparameters and specific \glspl{cnn}.

Following up upon the work \cite{pahde2023optimizing} that optimized \gls{lrp} \wrt specific explainability metrics, we argue that \gls{lrp} or any other tunable attribution method can be optimized specifically for the goal of more effective pruning (see \cref{fig:introduction}). 
In addition, recent works \cite{pmlr-v235-achtibat24a, ali2022xai} have extended \gls{lrp} to transformer-based architectures, allowing us now to also apply attribution-based pruning to the \gls{vit} architecture.

In this work, we revisit attribution-based model sparsification,
and extend the current state-of-the-art by
\begin{enumerate}
    \item proposing a novel pruning framework based on optimizing attribution method hyperparameters to achieve higher sparsification rates.
    \item incorporating transformer-based architectures, such as \glspl{vit}, to our framework by using recently developed attribution methods.
    \item discussing differences between \glspl{cnn} and \glspl{vit} in terms of pruning and over-parameterization.
    \item revealing that attributions optimized for explanations are not necessarily the best for pruning.
\end{enumerate}

\section{Related Work}
In the following,
we introduce related works in the field of \gls{xai}, efficient Deep Learning and the intersection between both.

\paragraph{Explainability and Local Feature Attribution Methods.}
Research in local explainability led to a plethora of methods (\eg, \cite{lundberg2017Shap, ribeiro2016should}),
commonly resulting in local feature attributions quantifying the importance of input features in the decision-making process.
These attributions are often shown in the form of heatmaps in the vision domain.
Notably, 
methods based on (modified) gradients, backpropagate attributions from the output to the input through the network,
conveniently offering attributions of \emph{all} latent components and neurons in a single backward pass \cite{sundararajan2017axiomatic, smilkov2017smoothgrad}.
Gradient-based attribution methods, however, can suffer from noisy gradients, rendering them unreliable for deep architectures \cite{balduzzi2017shattered}. 
Prominently,
\gls{lrp} \cite{bach2015pixel, montavon2019layer} introduces a set of different rules (with hyperparameters) that allow to reduce the noise level. 
In fact, as shown in \cite{pahde2023optimizing},
attribution methods such as \gls{lrp} can be optimized for certain \gls{xai} criteria, \eg, faithfulness or complexity~\cite{montavon2019layer, hedstrom2023quantus}.
We follow up on this observation, and specifically optimize \gls{xai} \wrt the task of pruning. That is, we add \gls{nn} pruning as an XAI evaluation criterion to optimize for.

\paragraph{Pruning of Deep Neural Networks.}
For pruning \glspl{cnn},
either individual (kernel) weights or whole structures, \eg, filters of convolution layers or neurons can be recognized as candidates for pruning \cite{he2018soft}.
For transformer architectures, such structures include heads of attention modules or linear layers inside transformer blocks \cite{voita2019analyzing, lagunas2021block}. 
In order to prune such structures, 
several criteria have been proposed to indicate which components are best suited to be removed, retaining performance as best as possible. 
The work of \cite{han2015learning} suggests pruning parameters based on weight magnitudes, offering a computationally free criterion.
Alternatively,
in \cite{dong2017activation}, the authors propose to prune neurons based on their activation patterns.
However, recent work~\cite{becking2020ecq} has shown that weight or activation magnitudes do not necessarily reflect a component's contribution during the inference process, \eg, also a small weight has the potential to be very relevant for the prediction of a class.

\paragraph{Explainability For Efficient Deep Learning.}
The work of~\cite{yeom2021pruning} introduces a novel pruning criterion based on \gls{xai},
and proposes to use latent relevance values extracted from the attribution method of \gls{lrp}~\cite{bach2015pixel, montavon2017explaining} to assign importance scores to network structures.
By taking into account how structures are used during the inference process, the work of \cite{yeom2021pruning} can effectively improve the efficiency-performance tradeoff.
Later,
the works of \cite{becking2020ecq,soroush2023compressing} also illustrate the value of \gls{xai} methods for network quantization.
In all these works,
heuristically chosen 
hyperparameters for \gls{lrp} have been used, 
which overlooks the potential for specific optimization to the task of pruning.
Notably,
\gls{lrp} is model-specific and thus restricted to compatible architectures.
In this work,
we also include recent \gls{lrp} extensions to transformer architectures,
and observe, that a specific \gls{lrp} rule commonly result in high pruning performances.

\section{Methods}
This work proposes a framework for pruning \glspl{dnn} using attribution methods from the field of \gls{xai} with hyperparameters specifically optimized for sparsification.
We begin with presenting our method in the form of a general \gls{xai}-based pruning principle in \cref{sec:attribution_pruning},
followed by introducing \gls{lrp} attributions and corresponding hyperparameters suitable for optimization, in \cref{sec:methods:lrp} and \cref{sec:methods:lrp_hyperparameters}, respectively. 
Lastly, \cref{sec:methods:hyperparameter_optimization} describes our optimization methodology.

\subsection{Attribution-based Pruning}
\label{sec:attribution_pruning}
For our structured pruning framework,
we view a \gls{dnn} as a collection of $p$ (interlinked) components $ \Psi = \{\psi_{1}, \dots, \psi_{p}\}$, that can correspond to, \eg, whole layers, (groups of) neurons, convolutional filters or attention heads.
We further assume access to an attribution method that generates attribution scores (relevance scores) $R_{\psi_{k}}({x_i})$ of component 
$\psi_{k} \in \Psi$
for the prediction of a sample $x_i$. 
The overall relevance of $\psi_{k}$ is then estimated through the mean relevance over a set of reference samples $\mathcal{X}_\text{ref} = \{x_1, x_2, \dots, x_{n_{\text{{ref}}}}\}$ as

\begin{figure}[t]
  \centering
    \includegraphics[width=0.89\linewidth]{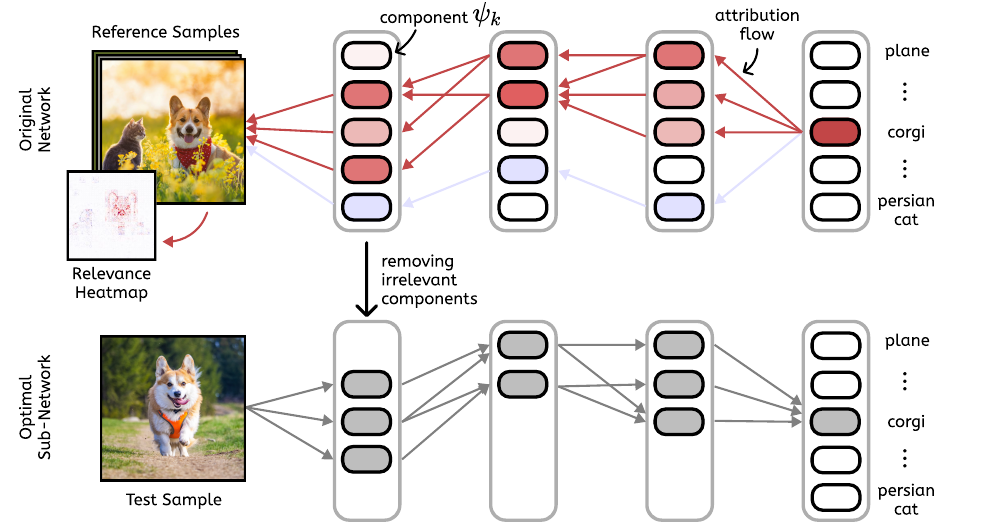}
   \caption{
   Attribution-based pruning workflow:
   Firstly,
   the relevant model structures are 
   identified by explaining a set of reference samples.
   The attribution method of choice (here \gls{lrp}) highlights the components and paths in the network which positively and negatively contribute to the decision-making.
   Positive and negative relevances are indicated by red and blue color respectively, and components with white color indicate zero or low relevance. Removing structures that receive the least relevance results in a sparser subnetwork, which performs significantly better than after random pruning.
   Notably,
   relevances can be computed \wrt a subset of output classes (\eg, ``corgi'' only),
   resulting in a subnetwork specifically designed to perform the restricted task. Credit: Nataba/iStock.
   }
   \label{fig:lrp-pruning}
\end{figure}

\begin{equation}
    \label{eq:compute_importance}
    \bar{R}_{\psi_{k}} = \frac{1}{n_{\text{ref}}}\sum_{i=1}^{n_{\text{ref}}} R_{\psi_{k}}({x_i})
\end{equation}

We collect relevance scores for all components via the set $\mathcal{R}$, given as
\begin{equation}
    \mathcal{R} = \{\bar{R}_{\psi_{1}}, \bar{R}_{\psi_{2}}, \dots, \bar{R}_{\psi_{p}}\}
\end{equation}
which, in turn, allows to define a pruning order for the model components.
Specifically, 
the indices $c$ for the components to be pruned up to the $q$-th place are given by

\begin{equation}
    \centering
    \{c\}_{q} = \text{argsort}(\mathcal{R})_{1, 2,\dots, q}\,
\end{equation}
resulting in the set of least relevant components.
Finally, the $q$ least relevant components are pruned by removing or masking the components from the computational graph as
\begin{equation}
    \centering
    \forall {\psi_{i} \in \Psi}: \psi_{i} \mapsto (1- \textbf{1}_{i \in \{c\}_{q}}) \psi_{i}
\end{equation}
where $\textbf{1}$ is an indicator function with condition $i \in \{c\}_{q}$.
The whole attribution-based pruning workflow is depicted in \cref{fig:lrp-pruning}.
Notably, previous work \cite{yeom2021pruning} demonstrates that \gls{lrp} is an effective method for attributing latent structures. \gls{lrp} further offers several hyperparameters to tune, which makes it a versatile choice for our framework.

\subsection{Layer-wise Relevance Propagation}
\label{sec:methods:lrp}

\glsdesc{lrp}~\cite{bach2015pixel, montavon2017explaining} is a rule-based backpropagation algorithm that was designed as a tool for interpreting non-linear learning models by assigning attribution scores, called ``relevances'', to network units proportionally to their contribution to the final prediction value.
Unlike other gradient- or perturbation-based methods, LRP treats a neural network as a layered directed acyclic graph with $L$ layers and input $x$:
\begin{equation}
    \centering
    f(x) = f^{L} \circ \dots \circ f^{l} \circ f^{l-1} \circ \dots \circ f^{1} (x)
\end{equation}
Beginning with an initial relevance score $R_j^L$ at output $j$ of layer $f^{L}$
(usually set as $f_{j}^{L}$ for an output of choice $j$), the score is layer-by-layer redistributed through all latent structures
to its input variables depending on the contribution from the these units to the output value:

Given a layer, we consider its pre-activations $z_{ij}$ mapping inputs $i$ to outputs $j$ and their aggregations $z_{j} = \sum_{i} z_{ij}$. 
Commonly in linear layers such a computation is given with  $z_{ij} = a_i w_{ij}$, where $w_{ij}$ are its weight parameters and $a_i$ the activation of neuron $i$.

Then, LRP distributes relevance quantities $R_j^l$ received from upper layers towards lower layers proportionally to the relative contributions of $z_{ij}$ to $z_j$, i.e., 
\begin{equation} \label{eq:lrp_basic}
    \centering
    R_{i\leftarrow j}^{(l-1, l)} = \frac{z_{ij}}{z_{j}} R_{j}^l
\end{equation}
In other words, the relevance message $ R_{i\leftarrow j}^{(l-1, l)}$ quantifies the contribution of neuron $i$ at layer $l-1$, to the activation of neuron $j$ at layer $l$.

To obtain the contribution of neuron $i$ to all upper layer neurons $j$, all incoming relevance messages $R_{i\leftarrow j}^{(l-1, l)}$ are losslessly aggregated as
\begin{equation} \label{eq:lrp_aggregation}
    R_i^{l-1} = \sum_j R_{i\leftarrow j}^{(l-1, l)}
\end{equation}

This process ensures relevance conservation between adjacent layers:

\begin{equation}
\label{lrp:conservation}
      \sum_i R^{l-1}_i = \sum_{i,j} R^{(l-1, l)}_{i\leftarrow j} = \sum_j R^l_j 
\end{equation}
which guarantees that the sum of all relevance in each layer stays the same.

When a group of neurons performs the same task as within a convolutional channel or attention head, it is beneficial to aggregate the total relevance of the entire group into a single relevance score. This aggregation process helps in simplifying the analysis and interpretation of the model's behavior by focusing on the collective relevance of each convolutional channel or attention head, rather than examining individual neurons. Further discussions of component-wise aggregation are given in \cref{app:aggregations}.

\subsection{Tuneable Hyperparameters of LRP} \label{sec:methods:lrp_hyperparameters}

Within the \gls{lrp} framework, various rules (as detailed in \cref{app:sec:lrp_rules}) have been proposed to tune relevance computation for higher explainability.

Further, several works have shown that applying different rules to variable parts of a network (referred to as a ``composite'') \cite{montavon2019layer, kohlbrenner2020towards, pmlr-v235-achtibat24a} enhances faithfulness and robustness of explanations.
We follow up on this idea, and perform a large-scale analysis over a variety of rules for different network parts. Inspired by the tuning configurations of \cite{pahde2023optimizing}, we split a network into four parts:

\begin{enumerate}
    \item the final classification layers, denoted as \gls{fcl},
    \item the last 25\% of hidden layers before the \gls{fcl}, denoted as \gls{hll},
    \item the first 25\% of hidden layers, denoted as \gls{lll},
    \item and the remaining hidden layers between \gls{hll} and \gls{lll}, denoted as \gls{mll}
\end{enumerate}

For each group, a specific rule set can be chosen, which together ultimately form the \gls{lrp} composite.
We refrain from treating each individual layer as a separate group due to the increased computational expenses associated with this setting during hyperparameter search.

For transformer architectures,
the works of \cite{pmlr-v235-achtibat24a, ali2022xai} proposed novel \gls{lrp} rules to attribute softmax non-linearities in the attention modules
(see \cref{app:sec:lrp_rules} for more details), encouraging us to also optimize \gls{lrp} configurations for the softmax non-linearities, resulting in an additional hyperparameter choices here.

It is important to note,
that relevance values can be positive or negative.
Components with positive relevance values indicate that they significantly contribute towards the explained class prediction.
Conversely, components with negative relevance values diminish the classification score, effectively speaking against the predicted class, as visible in \cref{fig:lrp-pruning}  where ``cat'' features (and the respective model components) speak against the ``corgi'' prediction. 
Components with near zero relevance do not contribute to the inference process in any meaningful way. 
Thus, a question arises \wrt the optimal order of pruning: Should we start with negative, or near zero relevant components?
To that end, we also add a hyperparameter indicating whether absolute relevance values are used as in \cref{eq:compute_importance}, \ie, $\bar{R}_{\psi_{k}} \mapsto |\bar{R}_{\psi_{k}}|$ for all $\psi_{k} \in \Psi$ if true.
The work of \cite{yeom2021pruning} takes into account and computes
only positive attributions by their \gls{lrp} rule choice (see \gls{lrp}-$z^{+}$ in \cref{app:sec:lrp_rules}). 
Our experiments in \cref{app:sec:optimized_composites}, however, reveal that the highest pruning rates are achieved by computing both positive and negative relevance signals, and then first pruning components with near zero relevance, as their minimal contribution in any direction
ensures low impact on the overall model performance.

\subsection{Hyperparameter Optimization Procedure} \label{sec:methods:hyperparameter_optimization}

In order to optimize the hyperparameters of an attribution method, we first define an optimization objective $C$.
As we perform our analysis for classification tasks,
we measure the top-1 accuracy on the validation dataset.
In principle,
a different performance criterion can be chosen here.
Concretely,
we measure model performance for different pruning rates $\text{PR}_{i} = \frac{1}{m} i$ as given after step $i\in \{0, \dots m-1\}$ of in total $m$ steps (excluding a 100\% rate).
After sequentially increasing the pruning rate,
and plotting performance against sparsity, we receive a curve that indicates the sparsity-accuracy tradeoff as, \eg, shown in \cref{fig:introduction} (\emph{left}).
We therefore propose to measure the resulting area under the curve $A_{\text{PR}}$ as given by

\begin{equation}
\label{eq:auc_cr}
    \centering
    A_{\text{PR}} = \frac{1}{m} \sum_{i=0}^{m-1} C(f_{\Psi_{\text{PR}_{i}}(\theta)})
\end{equation}

where the performance as given by $C$ depends on the network $f$ and its pruned parameters $\Psi_{\text{PR}_{i}}(\theta)$ after pruning step $i$ using the hyperparameter setting $\theta$ for the attribution method.
Ultimately, our objective will be the maximization of $A_{\text{PR}}$ \wrt $\theta$. 
Hyperparameter search is performed via grid search and Bayesian optimization techniques, as detailed in \cref{app:sec:optimization}.

\section{Experiments}
We begin our experiments with exploring the over-parameterization problem of \glspl{dnn} in \cref{experiments:overparam}.
This is followed by our results on finding the best \gls{lrp} hyperparameters for pruning \glspl{cnn} and \glspl{vit} in \cref{experiments:cnns} and \cref{experiments:vit}, respectively. Lastly,
we compare the effect of pruning using ideal and random attributions in \cref{experiments:behaviour} by evaluating how explanation heatmaps change after pruning.

\paragraph{Experimental Setting}
In our experiments,
we optimize \gls{lrp} hyperparameters for pruning convolution filters of VGG-16~\cite{simonyan2015very} (with and without BatchNorm~\cite{ioffe2015batch} layers),
ResNet-18 and ResNet-50~\cite{he2016deep} architectures (with 4224, 4224, 4800 and 26560 filters overall, respectively),
as well as linear layers and attention heads of the ViT-B-16 transformer~\cite{dosovitskiy2020image} (with 46080 neurons and 144 heads).
All models are pre-trained~\cite{marcel2010torchvision} and evaluated on the ImageNet dataset~\cite{deng2009imagenet}.
For hyperparameter optimization,
we measure model performance for 20 pruning rates (from 0\,\% up to 95\,\%) on the validation dataset.
To compute latent attributions,
a set of reference samples has been chosen from the training set of ImageNet (different set sizes are discussed later in \cref{experiments:cnns} and \ref{experiments:vit}).

\subsection{How Over-Parameterized are Vision Models?}
\label{experiments:overparam}

Training a large \gls{dnn} from scratch to solve a specific task can be computationally expensive.
A popular approach for saving training resources is to instead fine-tune a large pre-trained (foundation) model,
which often requires fewer training epochs and provides high (or even higher) model generalization, especially in sparse data settings~\cite{bommasani2021opportunities}.
Notably,
one of the effects that arise when solving for simple(r) tasks,
is that we likely end up with an over-parameterized model as only a subset of very specialized latent features are necessary to solve the task,
which makes pruning especially interesting in this case.

\begin{figure}[t]
  \centering
    \includegraphics[width=1\linewidth]{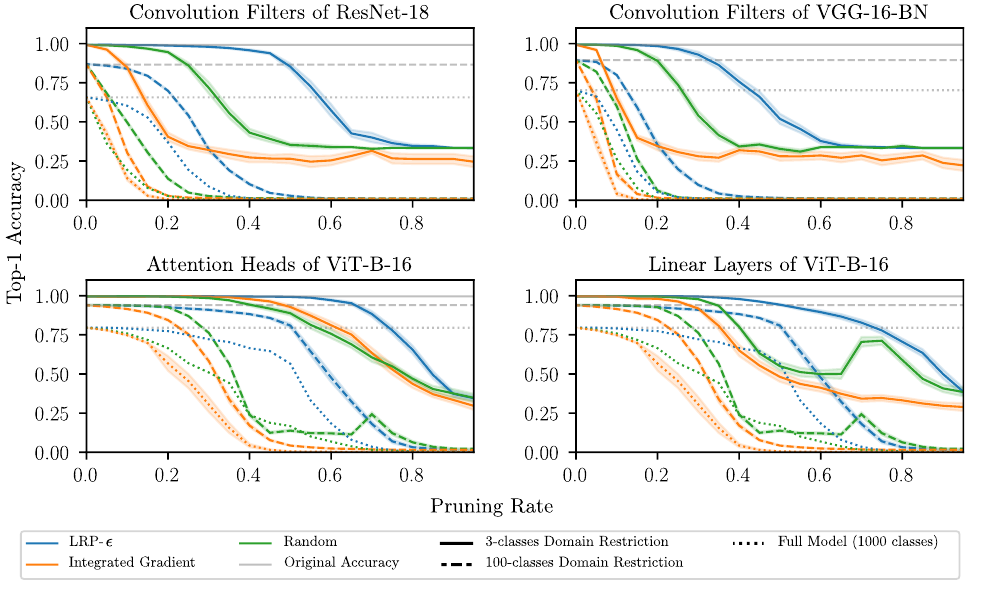}
  \caption{Investigating over-parameterization in \glspl{dnn} through attribution-based pruning (\gls{lrp}-opt in \emph{blue}, Integrated Gradient in \emph{orange color}) and random pruning (\emph{green color}). We compare pruning of all models \wrt different task difficulties, \ie, to differentiate between 1000 (\emph{dotted line}), 100 (\emph{dashed line}) or three ImageNet classes (\emph{solid line}).
  High performance for high sparsification rates indicates over-parameterization, \ie, many network components are not important for the task.
  Compared to the ResNet-18 and VGG-16-BN \glspl{cnn} (\emph{top}), the ViT-B-16 transformer shows a higher degree of over-parameterization (\emph{bottom}). \gls{sem} is illustrated (\emph{shaded area}) in the current and all other figures.
  }
   \label{fig:overparam}
\end{figure}

In fact,
when pruning the ResNet-18 and VGG-16-BN models that were pre-trained to detect \emph{all} of the 1000 ImageNet classes,
parameter count can only be reduced by a couple of percent without meaningful degradation of model performance, as shown in \cref{fig:overparam} (\emph{top}).
This indicates that ImageNet itself is a complex task, and most of the trained parameters are actually relevant.
However, as the task becomes simpler (simulated by reducing the number of output classes to 100 or three in the evaluation),
pruning rates can be increased to a much higher level without critical accuracy loss.

For baseline comparisons, we utilize random pruning, Integrated Gradient~\cite{sundararajan2017axiomatic} and our optimized LRP attribution (\gls{lrp}-opt) (see \cref{experiments:cnns}). We intentionally exclude naive weight magnitude or activation pruning since \cite{yeom2021pruning} has previously demonstrated that XAI-based methods outperform these traditional approaches by a large margin. Notably, especially for simpler tasks, attribution-based pruning outperforms random pruning, as attributions are output-specific and allow identifying the sub-graph that is relevant for the restricted task, as also illustrated in \cref{fig:lrp-pruning}.

Compared to the previously discussed \glspl{cnn},
the ViT-B-16 vision transformer seems to be more over-parameterized, as visible in \cref{fig:overparam} (\emph{bottom}), and therefore easier to prune.
Up to a pruning rate of 20\,\%, no meaningful accuracy loss is measured for detecting the 1000 ImageNet classes, irrespective of whether attention heads or neurons in linear layers are pruned.
This might be due to the fact,
that the vision transformer consists of about 46,000 neurons, which is approximately ten times the number of convolutional filters of the \glspl{cnn}.
Also, when comparing the ResNet-50 and ResNet-18 (see \cref{app:fig:overparam} for ResNet-50), the larger network shows higher pruning potential.

\subsection{Finding the Optimal Attributions for CNNs}
\label{experiments:cnns}

\begin{figure}[t]
  \centering
    \includegraphics[width=1\linewidth]{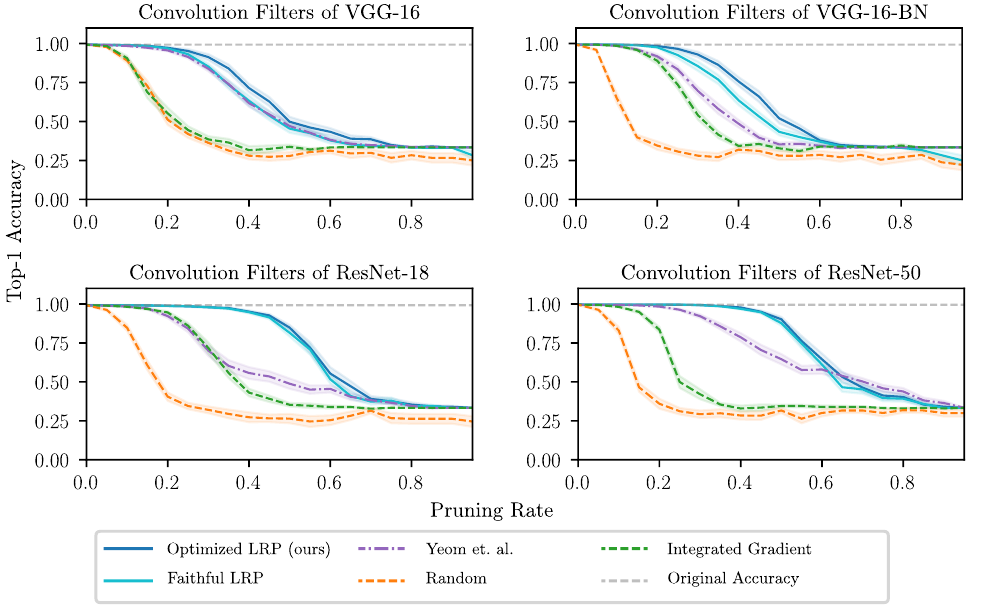}
  \caption{Pruning \glspl{cnn} models pre-trained on ImageNet (simplified task to detecting three classes), using ten reference samples per class. Results show a better sparsification-performance tradeoff for our optimized \gls{lrp} composite compared to a heuristic (faithful) \gls{lrp} composite, Yeom \etal \cite{yeom2021pruning} (details for each in \cref{app:tab:cnn_composites}, \cref{app:sec:faithful_lrp}, and \cref{eq:lrp-zplus}) and random pruning.}
   \label{fig:cnn_pruning}
\end{figure}

\begin{table*}[t]
    \centering
    \label{tab:cnn_pruning}
    \caption{Results for pruning \gls{cnn} models pre-trained on ImageNet. \textbf{Top-PR} demonstrates the highest pruning rate while keeping 95\% of baseline accuracy. \textbf{F}, \textbf{IG}, and \textbf{R} in order indicate a faithful \gls{lrp} composite, \text{Integrated Gradients}, and a \text{random} pruning baseline. Our results remark high scores for the area under the induced curve ($A^{\text{PR}}$) in pruning, overall improving the performance-sparsity tradeoff.}
    \begin{tabular}{@{}l@{\hspace{0.6em}}c@{\hspace{0.4em}}c@{\hspace{0.4em}}c@{\hspace{0.4em}}c@{\hspace{0.3em}}c@{\hspace{0.4em}}c@{\hspace{0.4em}}c@{\hspace{0.4em}}c@{\hspace{0.4em}}c@{\hspace{0.3em}}c}
        \\
        \toprule
        & \multicolumn{5}{c}{\textbf{{\text{$A^{\text{PR}}$}}}} & \multicolumn{5}{c}{\textbf{Top-PR ($\%$)}} $\pm$ 3 ($\%$) \\
         \cmidrule(lr){2-6} \cmidrule(lr){7-11}
        \textbf{Models} & \textbf{Ours} & \textbf{F} & \textbf{\cite{yeom2021pruning}} & \textbf{IG} & \textbf{R} & \textbf{Ours} & \textbf{F} & \textbf{\cite{yeom2021pruning}} & \textbf{IG} & \textbf{R} \\
        \midrule
        VGG-16        & \textbf{0.61}  & 0.58 & 0.58 & 0.43 & 0.4 & \textbf{26} & 23 & 21 & 7 & 6 \\ 
        VGG-16-BN     & \textbf{0.61} & 0.57 & 0.53 & 0.50 & 0.35 & \textbf{28} & 23 & 17 & 16 & 5 \\ 
        ResNet-18     & \textbf{0.69} & 0.68 & 0.57 & 0.53 & 0.37 & \textbf{41} & 40 & 18 & 20 & 5 \\ 
        ResNet-50     & \textbf{0.72} & 0.71 & 0.67 & 0.47 & 0.37 & \textbf{45} & \textbf{45} & 27 & 15 & 5 \\ 
        \bottomrule
    \end{tabular}
\end{table*}

Motivated by the previous \cref{experiments:overparam},
we in the following simulate a setting with over-parameterized networks that allows us to measure more significant differences between methods.
Specifically,
we restrict the data domain to three ImageNet classes and evaluate pruning using 20 different random seeds.
Experiments using toy models in the work of \cite{yeom2021pruning} showed that
ten (or more) reference samples are already well suited for estimating the overall relevances of network components (as used in \cref{eq:compute_importance}). 
We validated their finding also for the ResNet-18 model, as depicted in \cref{fig:num_ref_samples_resnet18}. 

Whereas we begin with optimizing \gls{lrp} for \gls{cnn} pruning,
we follow up with transformers in the next section.
Regarding \glspl{cnn}, we can strongly reduce the accuracy-sparsity tradeoff when pruning the convolution filters with our method based on optimizing \gls{lrp} attributions compared to other baselines, \eg, the heuristically chosen variant of \gls{lrp} used by \cite{yeom2021pruning} (see \cref{eq:lrp-zplus}), as also illustrated in \cref{fig:cnn_pruning} and \cref{tab:cnn_pruning}. 
Extensive hyperparameter search reveals that across different \gls{cnn} architectures,
simple hyperparameter settings exist that are well-suited for pruning in general.
One such candidate is the \gls{lrp}-$\epsilon$ rule (see  \cref{eq:lrp-epsilon}) applied for all convolutional layers. 
We refer to \cref{app:tab:cnn_composites} for a list of the best performing hyperparameter settings.
Interestingly for CNNs, 
\gls{lrp} composites known to lead to faithful explanations (in the suggested context of \cite{samek2016evaluating, bach2015pixel}) optimized by \cite{kohlbrenner2020towards}, are also effective in pruning.
One such composite (detailed in \cref{app:sec:faithful_lrp}) is also shown in \cref{fig:cnn_pruning} and denoted as ``Faithful \gls{lrp}''. 
Moreover, our results indicate a larger amount of unused structures across architectures with shortcut connections, \ie, the ResNet models.
As intermediate features can be passed through the short-cuts, while layers have the potential to be not used at all.

It is further not surprising that faithful \gls{lrp} composites \wrt \emph{input} explanations also perform well in pruning \emph{latent} structures.
Commonly, faithfulness is measured by ``deleting'' or perturbing input features and measuring the effect on the model's prediction~\cite{samek2016evaluating,hedstrom2023quantus}.
A highly (or lowly) relevant feature is assumed to result in a high (or low) change in model output.
As such,
attribution-based pruning in combination with measuring the model performance resembles a faithfulness evaluation scheme in \emph{latent} space.
The more faithful the attributions,
the smaller should be the degradation of model performance.
However,
later, in \cref{experiments:vit}, we can also observe ineffective pruning with an attributor that is known to be faithful in input space.

It is to note that the \gls{lrp}-$\varepsilon$ rule is known to result in noisy (and not necessarily faithful~\cite{samek2016evaluating}) input attributions for \glspl{dnn}, but performs very well for attributing latent components, as visible in \cref{app:tab:cnn_composites}.
We hypothesize that this results from fewer noise in intermediate layers~\cite{balduzzi2017shattered} and the fact that we aggregate attributions for each component (\eg, channel) which further reduces noise, as also proposed for gradients for the GradCAM method~\cite{selvaraju2017grad}.
High \emph{latent} faithfulness with noisy input attributors (\gls{lrp}-$\varepsilon$ or gradient times input~\cite{shrikumar2017learning}) has also been observed in \cite{dreyer2023revealing,fel2024holistic}.

\subsection{Finding the Optimal Attributions for Vision Transformer}
\label{experiments:vit}
\begin{figure}[t]
  \centering
    \includegraphics[width=1\linewidth]{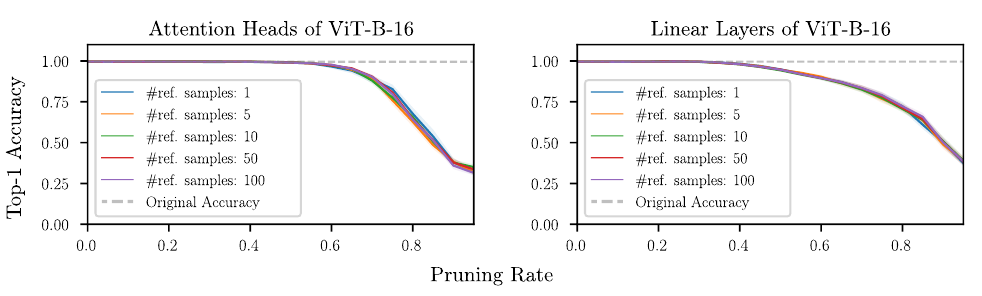}
  \caption{Attribution-based pruning using a different number of reference samples (per class) to estimate the importance of attention heads (\emph{left}) or neurons in linear layers (\emph{right}) of the ViT-B-16. 
  This experiment has been conducted for 20 different random seeds. For the propagation of \gls{lrp}, LRP-$\epsilon$ has been set as our parameter for all layers (\cref{sec:methods:lrp_hyperparameters}), and \wrt the attribution of softmax operations (\cref{app:sec:lrp_transformers}).}
   \label{fig:num_ref_samples}
\end{figure}

\begin{figure}[t]
  \centering
    \includegraphics[width=1\linewidth]{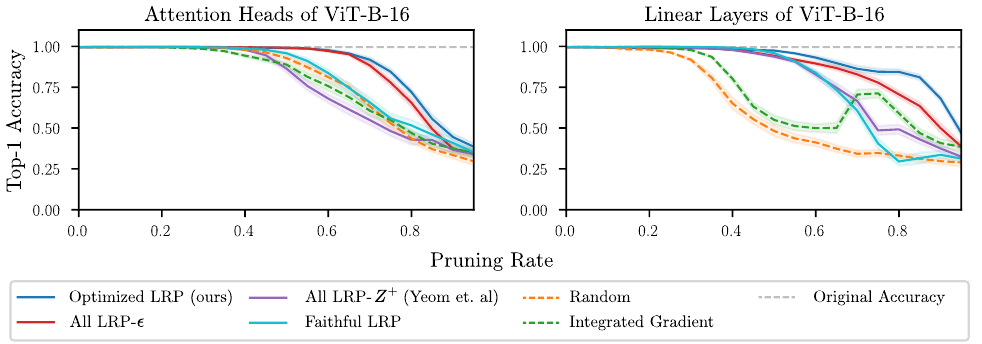}
  \caption{Pruning a \gls{vit} model pre-trained on ImageNet (simplified task to detecting three classes), using ten reference samples per class. Results show a better sparsification-performance tradeoff for our optimized \gls{lrp} composite compared to a heuristic (faithful) \gls{lrp} composite, Yeom \etal \cite{yeom2021pruning} (details for each in \cref{app:tab:cnn_composites}, \cref{app:sec:faithful_lrp}, and \cref{eq:lrp-zplus}, respectively), LRP-$\epsilon$, Integrated Gradient, and random pruning.}
   \label{fig:transformer_pruning}
\end{figure}

\begin{table*}[b]
\label{tab:transformer_pruning}
    \centering
    \caption{Results for pruning the ViT-B-16 pre-trained on ImageNet. \textbf{Top-PR} demonstrates the highest pruning rate while keeping 95\% of baseline accuracy. \textbf{F}, \textbf{IG}, and \textbf{R} in order indicate a faithful \gls{lrp} composite, \text{Integrated Gradients}, and a \text{random} pruning baseline. Our results remark high scores for the area under the induced curve ($A^{\text{PR}}$) in pruning, overall improving the performance-sparsity tradeoff.}
    \begin{tabular}{@{}l@{\hspace{0.6em}}c@{\hspace{0.4em}}c@{\hspace{0.4em}}c@{\hspace{0.4em}}c@{\hspace{0.3em}}c@{\hspace{0.4em}}c@{\hspace{0.4em}}c@{\hspace{0.4em}}c@{\hspace{0.4em}}c@{\hspace{0.4em}}c@{\hspace{0.4em}}c@{\hspace{0.3em}}c@{}}
        \\
        \toprule
        & \multicolumn{5}{c}{{{\text{$A^{\text{PR}}$}}}} & \multicolumn{5}{c}{\textbf{Top-PR ($\%$)} $\pm$ 3 ($\%$)} \\
         \cmidrule(lr){2-7} \cmidrule(lr){8-12}
        \textbf{Models} & \textbf{Ours} & \textbf{F} & \textbf{LRP-$\epsilon$} & \textbf{\cite{yeom2021pruning}} & \textbf{IG} & \textbf{R} & \textbf{Ours} & \textbf{F} & \textbf{LRP-$\epsilon$} & \textbf{\cite{yeom2021pruning}} & \textbf{IG} & \textbf{R} \\
        \midrule
        Linear Layers       & \textbf{0.87}  & 0.75 & 0.83 & 0.77 & 0.70 & 0.59 & \textbf{57} & 51  & 49  & 48 &  33 & 26 \\
        Attention Heads    & \textbf{0.85} & 0.78 & 0.85 & 0.74  & 0.75 & 0.76  & \textbf{66} & 51 & 64 & 44 & 39 & 47 \\
        
        \bottomrule
    \end{tabular}
\end{table*}

Attention heads and linear layers are two structures of interest in the pruning literature for Vision Transformers \cite{vaswani2017attention, lagunas2021block}. As observed in \cref{experiments:overparam}, 
the \gls{vit} model shows a higher degree of over-parameterization, possibly due to the abundance of neurons in the linear layers, ultimately highlighting the possible value for pruning. We now re-investigate the number of reference samples required to robustly estimate the relevance of a component in the ViT-B-16 model.
Interestingly, unlike CNNs, according to \cref{fig:num_ref_samples}, there is no deviation in pruning reliability and stability by using different numbers of reference samples.
Consequently, our experiments have been applied with exact same settings as in \cref{experiments:cnns}.

Our results from \cref{fig:transformer_pruning} again confirm an improvement of our optimized attribution-based pruning scheme (best \gls{lrp} composites shown in \cref{app:tab:vit_composite}) compared to the previous work of \cite{yeom2021pruning}. 
Adopting the generally reliable composite of LRP-$\epsilon$ obtained from \glspl{cnn} (as in \cref{fig:cnn_pruning} and \cref{app:tab:cnn_composites}) is again a promising option. 
Nonetheless, unlike for \glspl{cnn},
a recently proposed faithful \gls{lrp} composite~\cite{pmlr-v235-achtibat24a} designed for the ViT-B-16 model (detailed in \cref{app:sec:faithful_lrp}), is \emph{not} ideal for pruning. 
This finding, and our experiment in \cref{app:experiment:evaluate_attnlrp_lrppruner} show that optimizing an attributor for two different contexts of faithfulness (input or latent space/pruning) (\cref{sec:methods:lrp_hyperparameters}) does not necessarily lead to an attributor that attributes faithfully in both input and latent space.
The poorer performance of heuristically tuned \gls{lrp} from \cite{yeom2021pruning} (\cref{eq:lrp-zplus}) compared to random when pruning attention heads, underscores both the greater challenge of pruning this structure and the significance of our proposed optimization procedure.

\subsection{How Pruning Affects Model Explanations}
\label{experiments:behaviour}

In addition to keeping track of the model performance, the field of \gls{xai} encourages us in the following to investigate the model behavior by observing explanation heatmaps. 
Concretely,
we expect in the ideal pruning scheme,
that heatmaps change as late as possible when increasing the pruning rate.
This reflects that the task-relevant components are retained as long as possible.

\begin{figure}[t]
  \centering
    \includegraphics[width=0.95\linewidth]{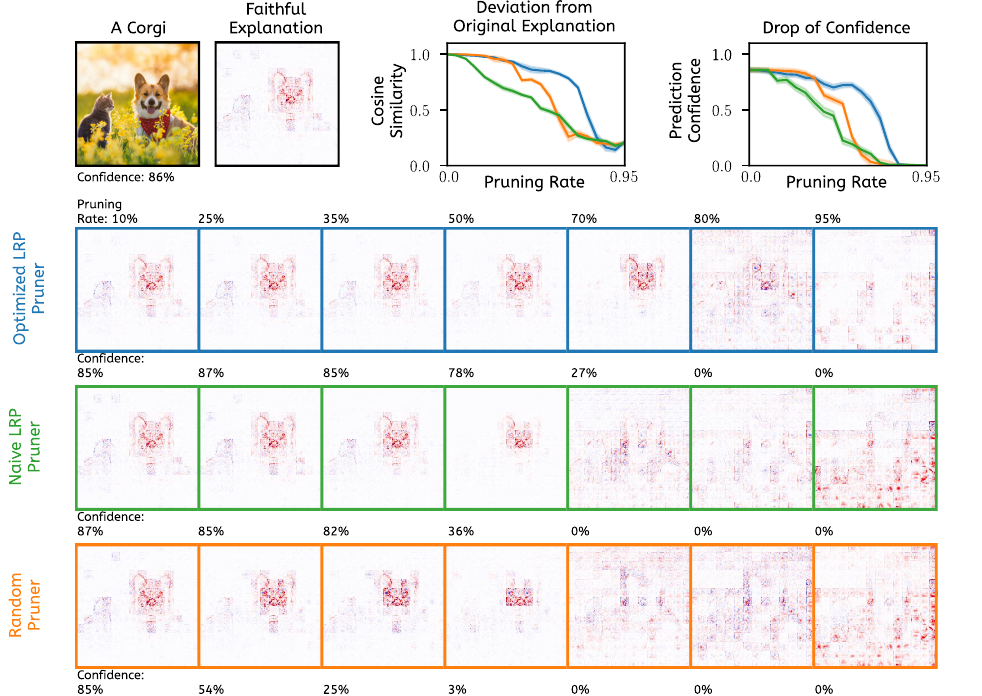}
  \caption{
  Pruning a \gls{vit} with the aim to retain a high accuracy for detecting ImageNet corgi classes (``Pembroke Welsh'' and ``Cardigan Welsh''). 
  We show, how explanation heatmaps for a corgi prediction change when increasing the pruning rate (\emph{bottom}). 
  Initially,
  most positive relevance lies on the corgi's head, with few negative relevance on a cat sitting next to the dog.
  As expected,
  when increasing the pruning rate using our optimized \gls{lrp} composite,
  the background features disappear before corgi-related features are perturbed.
  When naively using the composite proposed in \cite{yeom2021pruning}, or performing random pruning,
  heatmaps change much earlier (and more randomly), indicating model degradation.
  In fact,
  we can measure a high correlation of 0.99 between heatmap change and confidence loss (\emph{top right}).
  Surprisingly,
  the unoptimized \gls{lrp} composite performs even worse than random pruning,
  stressing the need to optimize attribution methods before applying them heuristically.
  }
   \label{corgi_cosine_similarity}
\end{figure}

As an illustrative example, we prune the attention heads of the ViT-B-16 model pre-trained on ImageNet with the aim to predict ImageNet corgi classes (``Pembroke Welsh'' and ``Cardigan Welsh'') as shown in \cref{corgi_cosine_similarity}. 
As a quantitative measure,
we compute the cosine similarity between the original heatmap (using the recently proposed composite of~\cite{achtibat2023attribution}) and the heatmap of the pruned model for different pruning rates over the validation set.

\clearpage
\clearpage

On the one hand,
we can see that random pruning or unoptimized attribution-based pruning leads to a much earlier change in heatmaps compared to our optimized \gls{lrp}-based approach.
The heatmaps indicate that irrelevant features are removed first as the pruning rate is increased (\eg, ``cat'' features in \cref{corgi_cosine_similarity} get disregarded before corgi-related features in the process). 
However, a random pruning approach might wrongly omit an important structure first, as heatmap changes of ``mouth'' and ``ear'' features can be seen in \cref{corgi_cosine_similarity}.
On the other hand, as can be expected, the change in heatmap similarity highly correlates with the model confidence, resulting in a Pearson correlation coefficient of 0.99.

\section{Conclusion}
In this work, we propose a general framework for post-hoc \gls{dnn} pruning that is based on using and optimizing attribution-based methods from the field of \glsdesc{xai} for more effective pruning. 
For our framework,
the method of \gls{lrp} is well-suited by offering several hyperparameters to tune attributions.
When applying our framework,
we can strongly reduce the performance-sparsity tradeoff of \glspl{cnn} and especially \glspl{vit} compared to previously established approaches.
Vision transformers are on the one hand more sensitive towards hyperparameters, and also show higher over-parameterization.
Overall,
using local \gls{xai} methods for pruning irrelevant model components demonstrates high potential in our experiments.

\section*{Acknowledgements}
This work was supported by the Federal Ministry of Education and Research (BMBF) as grant BIFOLD (01IS18025A, 01IS180371I); the German Research Foundation (DFG) as research unit DeSBi (KI-FOR 5363); the European Union’s Horizon Europe research and innovation programme (EU Horizon Europe) as grant TEMA (101093003); the European Union’s Horizon 2020 research and innovation programme (EU Horizon 2020) as grant iToBoS (965221).

\newpage

%
%
\bibliographystyle{splncs04}
\bibliography{main}

\begin{thebibliography}{10}
\providecommand{\url}[1]{\texttt{#1}}
\providecommand{\urlprefix}{URL }
\providecommand{\doi}[1]{https://doi.org/#1}

\bibitem{achtibat2023attribution}
Achtibat, R., Dreyer, M., Eisenbraun, I., Bosse, S., Wiegand, T., Samek, W.,
  Lapuschkin, S.: From attribution maps to human-understandable explanations
  through concept relevance propagation. Nature Machine Intelligence
  \textbf{5}(9),  1006--1019 (2023)

\bibitem{pmlr-v235-achtibat24a}
Achtibat, R., Hatefi, S.M.V., Dreyer, M., Jain, A., Wiegand, T., Lapuschkin,
  S., Samek, W.: {A}ttn{LRP}: Attention-aware layer-wise relevance propagation
  for transformers. In: Proceedings of the 41st International Conference on
  Machine Learning. Proceedings of Machine Learning Research, vol.~235, pp.
  135--168. PMLR (21--27 Jul 2024)

\bibitem{ali2022xai}
Ali, A., Schnake, T., Eberle, O., Montavon, G., M{\"u}ller, K.R., Wolf, L.: Xai
  for transformers: Better explanations through conservative propagation. In:
  International Conference on Machine Learning. pp. 435--451. PMLR (2022)

\bibitem{bach2015pixel}
Bach, S., Binder, A., Montavon, G., Klauschen, F., M{\"u}ller, K.R., Samek, W.:
  On pixel-wise explanations for non-linear classifier decisions by layer-wise
  relevance propagation. PloS one  \textbf{10}(7),  e0130140 (2015)

\bibitem{balduzzi2017shattered}
Balduzzi, D., Frean, M., Leary, L., Lewis, J., Ma, K.W.D., McWilliams, B.: The
  shattered gradients problem: If resnets are the answer, then what is the
  question? In: International Conference on Machine Learning. pp. 342--350.
  PMLR (2017)

\bibitem{becking2020ecq}
Becking, D., Dreyer, M., Samek, W., M{\"u}ller, K., Lapuschkin, S.: Ecq x:
  explainability-driven quantization for low-bit and sparse dnns. In:
  International Workshop on Extending Explainable AI Beyond Deep Models and
  Classifiers. pp. 271--296. Springer (2020)

\bibitem{blücher2024decoupling}
Blücher, S., Vielhaben, J., Strodthoff, N.: Decoupling pixel flipping and
  occlusion strategy for consistent xai benchmarks (2024)

\bibitem{bommasani2021opportunities}
Bommasani, R., Hudson, D.A., Adeli, E., Altman, R., Arora, S., von Arx, S.,
  Bernstein, M.S., Bohg, J., Bosselut, A., Brunskill, E., et~al.: On the
  opportunities and risks of foundation models. arXiv preprint arXiv:2108.07258
   (2021)

\bibitem{deng2009imagenet}
Deng, J., Dong, W., Socher, R., Li, L.J., Li, K., Fei-Fei, L.: Imagenet: A
  large-scale hierarchical image database. In: 2009 IEEE conference on computer
  vision and pattern recognition. pp. 248--255. Ieee (2009)

\bibitem{dong2017activation}
Dong, J., Zheng, H., Lian, L.: Activation-based weight significance criterion
  for pruning deep neural networks. In: Image and Graphics: 9th International
  Conference, ICIG 2017, Shanghai, China, September 13-15, 2017, Revised
  Selected Papers, Part II 9. pp. 62--73. Springer (2017)

\bibitem{dosovitskiy2020image}
Dosovitskiy, A., Beyer, L., Kolesnikov, A., Weissenborn, D., Zhai, X.,
  Unterthiner, T., Dehghani, M., Minderer, M., Heigold, G., Gelly, S., et~al.:
  An image is worth 16x16 words: Transformers for image recognition at scale.
  In: International Conference on Learning Representations (2020)

\bibitem{dreyer2023revealing}
Dreyer, M., Achtibat, R., Wiegand, T., Samek, W., Lapuschkin, S.: Revealing
  hidden context bias in segmentation and object detection through
  concept-specific explanations. In: Proceedings of the IEEE/CVF Conference on
  Computer Vision and Pattern Recognition Workshops. pp. 3828--3838 (2023)

\bibitem{fel2024holistic}
Fel, T., Boutin, V., B{\'e}thune, L., Cad{\`e}ne, R., Moayeri, M., And{\'e}ol,
  L., Chalvidal, M., Serre, T.: A holistic approach to unifying automatic
  concept extraction and concept importance estimation. Advances in Neural
  Information Processing Systems  \textbf{36} (2024)

\bibitem{frankle2018lottery}
Frankle, J., Carbin, M.: The lottery ticket hypothesis: Finding sparse,
  trainable neural networks. arXiv preprint arXiv:1803.03635  (2018)

\bibitem{frazier2018tutorial}
Frazier, P.I.: A tutorial on bayesian optimization. arXiv preprint
  arXiv:1807.02811  (2018)

\bibitem{geng2022pruning}
Geng, L., Niu, B.: Pruning convolutional neural networks via filter similarity
  analysis. Machine Learning  \textbf{111}(9),  3161--3180 (2022)

\bibitem{gholami2022survey}
Gholami, A., Kim, S., Dong, Z., Yao, Z., Mahoney, M.W., Keutzer, K.: A survey
  of quantization methods for efficient neural network inference. In: Low-Power
  Computer Vision, pp. 291--326. Chapman and Hall/CRC (2022)

\bibitem{han2015learning}
Han, S., Pool, J., Tran, J., Dally, W.: Learning both weights and connections
  for efficient neural network. Advances in neural information processing
  systems  \textbf{28} (2015)

\bibitem{he2016deep}
He, K., Zhang, X., Ren, S., Sun, J.: Deep residual learning for image
  recognition. In: Proceedings of the IEEE conference on computer vision and
  pattern recognition. pp. 770--778 (2016)

\bibitem{he2018soft}
He, Y., Kang, G., Dong, X., Fu, Y., Yang, Y.: Soft filter pruning for
  accelerating deep convolutional neural networks. In: IJCAI International
  Joint Conference on Artificial Intelligence (2018)

\bibitem{hedstrom2023quantus}
Hedstr{\"{o}}m, A., Weber, L., Krakowczyk, D., Bareeva, D., Motzkus, F., Samek,
  W., Lapuschkin, S., H{\"{o}}hne, M.M.M.: Quantus: An explainable ai toolkit
  for responsible evaluation of neural network explanations and beyond. Journal
  of Machine Learning Research  \textbf{24}(34),  1--11 (2023)

\bibitem{howard2017mobilenets}
Howard, A.G., Zhu, M., Chen, B., Kalenichenko, D., Wang, W., Weyand, T.,
  Andreetto, M., Adam, H.: Mobilenets: Efficient convolutional neural networks
  for mobile vision applications. arXiv preprint arXiv:1704.04861  (2017)

\bibitem{ioffe2015batch}
Ioffe, S., Szegedy, C.: Batch normalization: Accelerating deep network training
  by reducing internal covariate shift. In: International conference on machine
  learning. pp. 448--456. pmlr (2015)

\bibitem{kohlbrenner2020towards}
Kohlbrenner, M., Bauer, A., Nakajima, S., Binder, A., Samek, W., Lapuschkin,
  S.: Towards best practice in explaining neural network decisions with lrp.
  In: 2020 International Joint Conference on Neural Networks (IJCNN). pp.~1--7.
  IEEE (2020)

\bibitem{kuzmin2024pruning}
Kuzmin, A., Nagel, M., Van~Baalen, M., Behboodi, A., Blankevoort, T.: Pruning
  vs quantization: which is better? Advances in neural information processing
  systems  \textbf{36} (2024)

\bibitem{lagunas2021block}
Lagunas, F., Charlaix, E., Sanh, V., Rush, A.M.: Block pruning for faster
  transformers. arXiv preprint arXiv:2109.04838  (2021)

\bibitem{lee2021layer}
Lee, J., Park, S., Mo, S., Ahn, S., Shin, J.: Layer-adaptive sparsity for the
  magnitude-based pruning. In: 9th International Conference on Learning
  Representations, ICLR 2021 (2021)

\bibitem{li2022efficientformer}
Li, Y., Yuan, G., Wen, Y., Hu, J., Evangelidis, G., Tulyakov, S., Wang, Y.,
  Ren, J.: Efficientformer: Vision transformers at mobilenet speed. Advances in
  Neural Information Processing Systems  \textbf{35},  12934--12949 (2022)

\bibitem{lundberg2017Shap}
Lundberg, S.M., Lee, S.: A unified approach to interpreting model predictions.
  In: Advances in Neural Information Processing Systems 30. pp. 4765--4774
  (2017)

\bibitem{marcel2010torchvision}
Marcel, S., Rodriguez, Y.: Torchvision the machine-vision package of torch. In:
  Proceedings of the 18th ACM international conference on Multimedia. pp.
  1485--1488 (2010)

\bibitem{molchanov2016pruning}
Molchanov, P., Tyree, S., Karras, T., Aila, T., Kautz, J.: Pruning
  convolutional neural networks for resource efficient inference. arXiv
  preprint arXiv:1611.06440  (2016)

\bibitem{montavon2019layer}
Montavon, G., Binder, A., Lapuschkin, S., Samek, W., M{\"u}ller, K.R.:
  Layer-wise relevance propagation: an overview. Explainable AI: interpreting,
  explaining and visualizing deep learning pp. 193--209 (2019)

\bibitem{montavon2017explaining}
Montavon, G., Lapuschkin, S., Binder, A., Samek, W., M{\"u}ller, K.R.:
  Explaining nonlinear classification decisions with deep taylor decomposition.
  Pattern recognition  \textbf{65},  211--222 (2017)

\bibitem{pahde2023optimizing}
Pahde, F., Yolcu, G.{\"U}., Binder, A., Samek, W., Lapuschkin, S.: Optimizing
  explanations by network canonization and hyperparameter search. In:
  Proceedings of the IEEE/CVF Conference on Computer Vision and Pattern
  Recognition. pp. 3818--3827 (2023)

\bibitem{peste2021ac}
Peste, A., Iofinova, E., Vladu, A., Alistarh, D.: Ac/dc: Alternating
  compressed/decompressed training of deep neural networks. Advances in neural
  information processing systems  \textbf{34},  8557--8570 (2021)

\bibitem{ribeiro2016should}
Ribeiro, M.T., Singh, S., Guestrin, C.: " why should i trust you?" explaining
  the predictions of any classifier. In: Proceedings of the 22nd ACM SIGKDD
  international conference on knowledge discovery and data mining. pp.
  1135--1144 (2016)

\bibitem{samek2016evaluating}
Samek, W., Binder, A., Montavon, G., Lapuschkin, S., M{\"u}ller, K.R.:
  Evaluating the visualization of what a deep neural network has learned. IEEE
  Transactions on Neural Networks and Learning Systems  \textbf{28}(11),
  2660--2673 (2017)

\bibitem{selvaraju2017grad}
Selvaraju, R.R., Cogswell, M., Das, A., Vedantam, R., Parikh, D., Batra, D.:
  Grad-cam: Visual explanations from deep networks via gradient-based
  localization. In: Proceedings of the IEEE international conference on
  computer vision. pp. 618--626 (2017)

\bibitem{shrikumar2017learning}
Shrikumar, A., Greenside, P., Kundaje, A.: Learning important features through
  propagating activation differences. In: International Conference on Machine
  Learning. pp. 3145--3153. PMLR (2017)

\bibitem{simonyan2015very}
Simonyan, K., Zisserman, A.: Very deep convolutional networks for large-scale
  image recognition. In: 3rd International Conference on Learning
  Representations (ICLR 2015). Computational and Biological Learning Society
  (2015)

\bibitem{smilkov2017smoothgrad}
Smilkov, D., Thorat, N., Kim, B., Vi{\'e}gas, F., Wattenberg, M.: Smoothgrad:
  removing noise by adding noise. arXiv preprint arXiv:1706.03825  (2017)

\bibitem{soroush2023compressing}
Soroush, K., Raji, M., Ghavami, B.: Compressing deep neural networks using
  explainable ai. In: 2023 13th International Conference on Computer and
  Knowledge Engineering (ICCKE). pp. 636--641. IEEE (2023)

\bibitem{sundararajan2017axiomatic}
Sundararajan, M., Taly, A., Yan, Q.: Axiomatic attribution for deep networks.
  In: International Conference on Machine Learning. pp. 3319--3328. PMLR (2017)

\bibitem{vaswani2017attention}
Vaswani, A., Shazeer, N., Parmar, N., Uszkoreit, J., Jones, L., Gomez, A.N.,
  Kaiser, {\L}., Polosukhin, I.: Attention is all you need. Advances in Neural
  Information Processing Systems  \textbf{30} (2017)

\bibitem{voita2019analyzing}
Voita, E., Talbot, D., Moiseev, F., Sennrich, R., Titov, I.: Analyzing
  multi-head self-attention: Specialized heads do the heavy lifting, the rest
  can be pruned. arXiv preprint arXiv:1905.09418  (2019)

\bibitem{williams1995gaussian}
Williams, C., Rasmussen, C.: Gaussian processes for regression. Advances in
  neural information processing systems  \textbf{8} (1995)

\bibitem{yang2022channel}
Yang, C., Liu, H.: Channel pruning based on convolutional neural network
  sensitivity. Neurocomputing  \textbf{507},  97--106 (2022)

\bibitem{yeom2021pruning}
Yeom, S.K., Seegerer, P., Lapuschkin, S., Binder, A., Wiedemann, S.,
  M{\"u}ller, K.R., Samek, W.: Pruning by explaining: A novel criterion for
  deep neural network pruning. Pattern Recognition  \textbf{115},  107899
  (2021)

\bibitem{zhou2016dorefa}
Zhou, S., Wu, Y., Ni, Z., Zhou, X., Wen, H., Zou, Y.: Dorefa-net: Training low
  bitwidth convolutional neural networks with low bitwidth gradients. arXiv
  preprint arXiv:1606.06160  (2016)

\end{thebibliography}


\newpage

\appendix

\renewcommand\thefigure{A.\arabic{figure}}    
\renewcommand\thetable{A.\arabic{table}}
\setcounter{figure}{0}
\setcounter{table}{0}
\renewcommand\theequation{A.\arabic{equation}}
\setcounter{equation}{0}
\section*{Appendix}

\section{Attribution Methods}

As it has been discussed in \cref{sec:attribution_pruning}, backpropagation methods are suitable for our proposed framework. We focus mostly on \gls{lrp} and it will be compared with Integrated Gradient as a commonly used gradient-based explainer.

\subsection{Integrated Gradient}
\label{a[[:sec:integrated_gradient]]}

As an improvement to pure gradient, Integrated Gradient \cite{sundararajan2017axiomatic} interpolates input $x$ for $m$ steps and computes their gradients sequentially.

\begin{align}
    \label{eq:integratedgrad}
    \text{IG}(x) &= (x - x^{\prime}) \int_{\alpha = 0}^{1} \frac{\partial f_{j}(x^{\prime} + \alpha \times (x - x^{\prime}))}{\partial x} dx \\
    &\approx (x - x^{\prime}) \sum_{k=1}^{m} \frac{\partial f_{j}(x^{\prime} + \frac{k}{m} \times (x - x^{\prime}))}{\partial x} \times \frac{1}{m}
\end{align}

Unlike \gls{lrp}, there is no parameter to be tuned in this method. $m$ from equation \cref{eq:integratedgrad} only serves as an approximation factor, with higher steps indicating more precise computation of the integral. For the experiments conducted in this paper, we interpolate each input for $20$ steps.

\subsection{Variants of LRP}
\label{app:sec:lrp_rules}

In addition to the vanilla rule of \gls{lrp} demonstrated in \ref{eq:lrp_basic}, \cite{bach2015pixel, montavon2019layer} later on proposed different rules as an extension of vanilla rule designed to serve different purposes. Most common extension of \gls{lrp} will be discussed based on the layer type they target.

\subsection{Tackling Linear and Convolution Layers}

As a side note, in the propagation process of \gls{lrp}, linear layers of \glspl{dnn} such as fully connected, and convolution layers, are treated similarly.

\subsubsection{LRP-$\epsilon$}
The most fundamental problem of rule \ref{eq:lrp_basic} causing computational instability, is division by zero which takes place when a neuron is not activated at all ($z_j=0$). LRP-$\epsilon$ with $\epsilon \in \mathbb{R}$ as a stabilizer parameter, was proposed to tackle this problem so that the denominator never reaches zero:

\begin{equation}
\label{eq:lrp-epsilon}
    \centering
    R_{i\leftarrow j} = \frac{z_{ij}}{z_{j} + \epsilon \text{sign}(z_j)} R_{j}
\end{equation}

As a side note, sign(0) = 1. Although the value of $\epsilon$ can be tuned, we set it to $1e-6$ in every use-case.

\subsubsection{LRP-$\alpha \beta$}
As described in figure \cref{fig:lrp-pruning} and \cref{sec:methods:lrp_hyperparameters}, relevances either have a positive or negative sign, indicating whether they have contributed positively or negatively in favor of the decision-making. Depending on the use-case, we might be interested in having a higher emphasis on positive relevances rather than the negative ones or vice versa. LRP-$\alpha \beta$ rule with $\alpha+\beta=1$, lets us apply modifications in that regard.

\begin{equation}
\label{eq:lrp-alphabeta}
    \centering
    R_{i\leftarrow j} = \bigr(\alpha \frac{z^{+}_{ij}}{z^{+}_{j}} + \beta \frac{z^{-}_{ij}}{z^{-}_{j}}\bigr) R_{j}
\end{equation}

\subsubsection{LRP-$z^{+}$}
\label{eq:lrp-zplus}
An extreme case for the above formulation when it is required to ignore all negative relevance, is the combination of $\alpha=1$ and $\beta=0$, which is referred to as LRP-$z^{+}$ used by \cite{yeom2021pruning}.

\subsubsection{LRP-$\gamma$} Another derivation, with the same purpose of LRP-$\alpha \beta$ to control over the distribution of positive and negative relevance separately, has this form:

\begin{equation}
\label{eq:lrp-gamma}
    \centering
    R_{i\leftarrow j} = \frac{z_{ij} + \gamma z_{ij}^+}
    {z_j + \gamma \sum_k z_{kj}^+} R_{j}
\end{equation}

Evidently, in case of $\gamma = \infty$, LRP-$\gamma$ will be equivalent to LRP-$z^{+}$.
The value of $\gamma$ can also be tuned in the optimization process. However, to prevent further computational costs unlike results from \cite{pmlr-v235-achtibat24a} in \cref{app:sec:faithful_lrp}, we refrain from tuning it.

\subsection{Tackling Non-linearities}
\subsubsection{Activation Functions}

In \gls{lrp} framework, a non-linear activation function $\sigma$ (\eg, ReLU), with $\sigma_{i}(z_i) = a_{i}$ is handled similar to \cref{eq:lrp_basic}, where $z_{ij} = \delta_{ij} z_{i}$ and $\delta$ represents the Kronecker delta.

\subsubsection{Attention and Softmax on Transformers}
\label{app:sec:lrp_transformers}
The key component of the attention module \cite{vaswani2017attention}, consists of two equations:
\begin{align}
    & \textbf{A} = \text{softmax}\left(\frac{\textbf{Q} \cdot \textbf{K}^T}{\sqrt{d_k}}\right) \label{eq:attention_softmax}\\
    & \textbf{O} = \textbf{A} \cdot \textbf{V} \label{eq:attention_multiplication}
\end{align}
where ($\cdot$) denotes matrix multiplication, $\textbf{K} \in \mathbb{R}^{b \times s_k \times d_k}$ is the key matrix, $\textbf{Q} \in \mathbb{R}^{b \times s_q \times d_k}$ is the queries matrix, $\textbf{V} \in \mathbb{R}^{b \times s_k \times d_v}$ the values matrix, and $\textbf{O} \in \mathbb{R}^{b \times s_k \times d_v}$ is the final output of the attention mechanism.
$b$ is the batch dimension including the number of heads, and $d_k, d_v$ indicate the embedding dimensions, and $s_q, s_k$ are the number of query and key/value tokens.

\subsubsection{CP-LRP}
\label{app:cp_lrp}
In the extension of \gls{lrp} from \cite{ali2022xai}, it was proposed to regard the attention matrix (softmax output) \textbf{A} in \cref{eq:attention_multiplication} as constant, attributing relevance solely through the value path by stopping the relevance flow through the softmax. Consequently, the matrix multiplication \cref{eq:attention_multiplication} can be treated with the $\varepsilon$-rule \cref{eq:lrp-epsilon}.

\subsubsection{AttnLRP}
The recent work of \cite{pmlr-v235-achtibat24a} extends LRP to Large Language Models and Vision Transformers by proposing to linearize the softmax function in \cref{eq:attention_softmax} with a Deep Taylor Decomposition~\cite{montavon2017explaining} which leads to substantially improved attributions in the natural language domain.
For that, the softmax non-linearity \cref{eq:attention_softmax} is attributed as follows:

\begin{equation}
    \label{eq:attnlrp_softmax}
    R^l_i = x_i (R^{l+1}_i - s_i \sum_j R^{l+1}_j)
\end{equation}
where $s_i$ denotes the $i$-th output of the softmax.
In addition, the matrix multiplication \cref{eq:attention_multiplication} is attributed with
\begin{equation}
    \label{eq:attnlrp_matrix_multiplication}
    R_{ji}^{l-1}(\textbf{A}_{ji}) = \sum_p \textbf{A}_{ji} \textbf{V}_{ip} \frac{R^l_{jp}}{2 \ \textbf{O}_{jp} + \epsilon}
\end{equation}
Finally, in order to mitigate noise in the attributions for Vision Transformers, the authors \cite{pmlr-v235-achtibat24a} propose to apply the LRP-$z^{+}$ on the Deep Taylor linearization of the softmax non-linearity \cref{eq:attention_softmax}:

\begin{equation}
    R_i^{l-1} = \sum_j (\textbf{J}_{ji} \ x_i)^+ \frac{R_j^l}{\sum_k (\textbf{J}_{jk} \ x_k)^+ + \tilde{b}_j^+}
\end{equation}
where $J$ is the Jacobian of the softmax computed at the current point $x$ and $\tilde{b}$ is the bias term of the linearization. For more details, please refer to \cite{pmlr-v235-achtibat24a}.

\subsection{Faithful LRP Composites}
\label{app:sec:faithful_lrp}

The works of \cite{kohlbrenner2020towards, pmlr-v235-achtibat24a} introduced 2 different \gls{lrp} composites as faithful explainer \cite{samek2016evaluating}. For CNNs, \cite{kohlbrenner2020towards} suggested applying LRP-$z^{+}$ to \glspl{fcl} and LRP-$\epsilon$ to other remaining layers (including \glspl{hll}, \glspl{mll}, \glspl{lll}).

On the other hand, experiments from \cite{pmlr-v235-achtibat24a} over the layers of transformers (differentiated based on layer types) demonstrated simple LRP-$\epsilon$ (\cref{eq:lrp-epsilon}) generates faithful explanation for \glspl{llm}. However, unlike \glspl{llm}, the composite of \cref{app:tab:vit_composite} seems to be more suitable for \glspl{vit}.

\begin{table*}[h!]
    \centering
    \caption{Faithful \gls{lrp} composite for Vision Transformers. LinearInputProjection and LinearOutputProjection are the linear layers generating $Q$, $K$, $V$, and $O$ inside attention-module of \cref{eq:attention_multiplication} and \cref{eq:attention_softmax}. The values of $\gamma$ for applying \gls{lrp} in Convolution and Linear layers are $0.25$ and, $0.05$ respectively.}
    \label{app:tab:vit_composite}
    \begin{tabular}{ccccc}
        \\
        \toprule
        Conv. & Linear & LinearInputProjection & LinearOutputProjection & Softmax\\
        \midrule
        $\gamma$ & $\gamma$ & $\epsilon$ & $\epsilon$ & Attn-LRP$(z^{+})$\\
        \bottomrule
    \end{tabular}
\end{table*}

\section{Spatial Dimension of To-Be-Pruned Components}
\label{app:aggregations}

In \cref{eq:compute_importance}, $R_{p}(x_{i})$ presented the relevance of component $p$ for reference sample $x_{i}$. However, it is crucial to discuss the spatial dimensionality of each component depending on its type because the relevance of a component gets aggregated over its dimensions.
Convolution filters of \glspl{cnn} have the size of the tuple $(h, w, d)$ indicating the height, weight, and depth of a filter. Thus, for a batch of size $n$ and convolution filter $p$, \cref{app:tab:vit_composite} will be transformed to:

\begin{equation}
    \centering
    \bar{R_{p}} = \sum_{i}^{n} \sum_{j}^{h} \sum_{l}^{w} \sum_{r}^{d} R_{(p, j, l, r)}({x_i})
\end{equation}

Unlike linear layers of \glspl{cnn} which simply follow \cref{eq:compute_importance}, an extra token axes $(t)$ for neurons inside linear layers of transformers resulting in this modification:

\begin{equation}
    \centering
    \bar{R_{p}} = \sum_{i}^{n} \sum_{j}^{t} R_{(p, j)}({x_i})
\end{equation}

Attention heads of \cref{eq:attention_softmax}, have two extra axes $(d_q, d_k)$ regarding as Query, and Key. Aggregation of relevance follows below formula:

\begin{equation}
    \centering
    \bar{R_{p}} = \sum_{i}^{n} \sum_{j}^{d_q} \sum_{l}^{d_k} R_{(p, j, l)}({x_i})
\end{equation}

However, in cases of computing the magnitude of relevance, aggregation should follow this structure:

\begin{equation}
    \centering
    \bar{R_{p}} = \sum_{i}^{n} \sum_{j}^{d_q} \big| \sum_{l}^{d_k} R_{(p, j, l)}({x_i})\big| 
\end{equation}

\section{Additionally on Overparameterization}

\begin{figure}[t]
  \centering
    \includegraphics[width=1\linewidth]{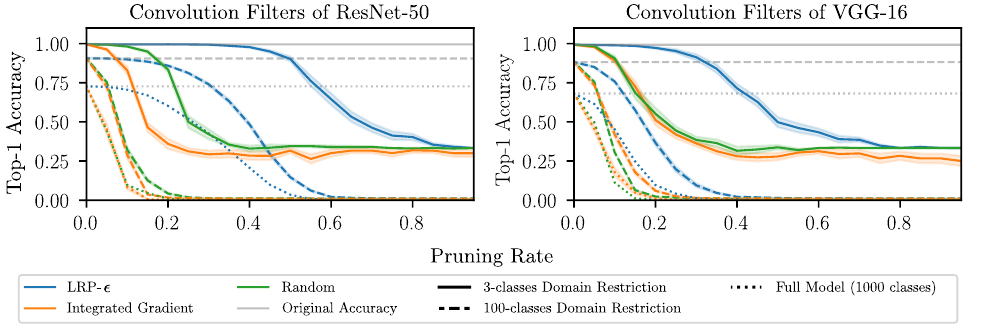}
  \caption{With settings similar to \cref{fig:overparam}, few-shot pruning demonstrates overparameterization exceeds, especially in cases of simpler tasks. Additionally, \textit{Relevance} denoted in the legend is computed by LRP framework with $\epsilon$ rule (see \cref{app:sec:lrp_rules}) applied across all layers.}
   \label{app:fig:overparam}
\end{figure}

Similar to \cref{fig:overparam}, \cref{app:fig:overparam} demonstrates the effect of overparameterization additionally on VGG-16 and ResNet-50 signifying again over-use of parameters in simpler tasks. Similarly in this case, architectures with skip connections show more potential to be overparameterized.

\section{On Optimization: From Bayesian to Grid Search}
\label{app:sec:optimization}

To find an optimal \gls{lrp} composite for pruning, our approach takes place by discovering prospective parameters (\cref{sec:methods:lrp_hyperparameters}) via Bayesian Optimization \cite{frazier2018tutorial} (with Gaussian Process regressor \cite{williams1995gaussian} as a surrogate model back-bone), followed by  Grid Search on the reduced parameter space. This seems to be a more effective solution rather than naively applying Grid Search over the whole parameters when we can reduce from $50\%$ up to $90\%$ of our search space approximately.

\section{On Number of Reference Samples}

\begin{figure}[t]
  \centering
    \includegraphics[width=1\linewidth]{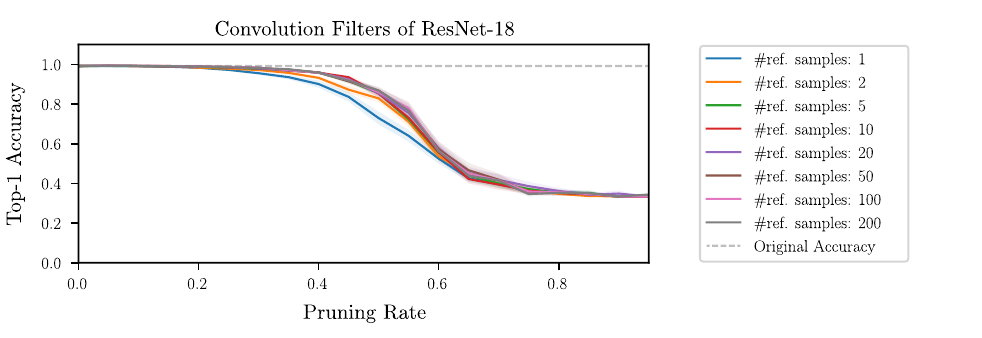}
   \caption{Tracking changes of pruning rate for a ResNet-18 pre-trained on ImageNet based on the variable number of reference samples to conduct pruning via \gls{lrp} for 20 different random seeds. We used LRP-$\epsilon$ over all layers.}
   \label{fig:num_ref_samples_resnet18}
\end{figure}

\cref{fig:num_ref_samples_resnet18} indicates and validates the findings of work \cite{yeom2021pruning} that using minimally 10 reference samples results in stable pruning for CNNs, as no more impact can be observed when using more samples. The fewer used samples per class lead to a lower pruning rate.

\section{Optimized Composites}
\label{app:sec:optimized_composites}

\subsection{Top Composites for Pruning}
In later tables, based on the conducted optimization, we demonstrate top composites for pruning on \glspl{cnn} (\cref{app:tab:cnn_composites}), ViT-B-16 (\cref{app:tab:transformer_composites}) and propose a general composite with persuasive performance to use over each architecture type individually. Later, in \cref{app:tab:ablation} we discuss which parameters are more important to be tuned.

\begin{table*}[h!]
\label{app:tab:cnn_composites}
    \centering
    \caption{Top composites per architecture for Pruning Convolution Filters of CNNs. The last column demonstrates the highest compression rate while keeping 95$\%$ of baseline accuracy (with $\pm$ 3$\%$ of approximation error). Last row indicates using LRP-$\epsilon$ (with $\epsilon=1e-6$) is a generally strong indicator for pruning these CNNs. Based on section \cref{sec:methods:lrp_hyperparameters}, the True flag of \gls{mag} suggests that sorting the importance of components should be based on the magnitude of the relevance rather than its sign.}
    \begin{tabular}{@{}l@{\hspace{0.6em}}c@{\hspace{0.4em}}c@{\hspace{0.4em}}c@{\hspace{0.4em}}c@{\hspace{0.3em}}c@{\hspace{0.5em}}c@{\hspace{0.5em}}c@{}}
        \\
        \toprule
        \textbf{Models}  & \textbf{LLL} & {\textbf{MLL}} & {\textbf{HLL}} & {\textbf{FC}} & {\textbf{\gls{mag}}} &($\uparrow$){\textbf{$A^{PR}$}} & Top-PR ($\%$)  \\
        \midrule

        VGG-16 & $\epsilon$ & $\epsilon$ & $\epsilon$ & $\epsilon$ & True & 0.608 & 26\\ 
        VGG-16-BN & $\epsilon$ & $\epsilon$ & $\epsilon$ & $\epsilon$ & True & 0.609 & 28 \\
        ResNet-18 & $\epsilon$ & $\epsilon$ & $\epsilon$ & $\alpha^{2}\beta^{1}$ & True & 0.691 & 41 \\
        ResNet-50 &  $\epsilon$ & $\epsilon$ & $\epsilon$ & $\epsilon$ & True & 0.718 & 45 \\
        \midrule
        Across All &  $\epsilon$ & $\epsilon$ & $\epsilon$ & $\epsilon$ & True & 0.656 &  \\
        \bottomrule
    \end{tabular}
\end{table*}

\begin{table*}[h!]
\label{app:tab:transformer_composites}
    \centering
    \caption{Top composites for pruning Linear Layers and Attention Heads of ViT-B-16. Similar to the previous table, \textbf{Top-PR} demonstrates the highest pruning rate while keeping 95\% of baseline accuracy (with $\pm$ 3$\%$ of approximation error). The last row also proposes a composite suitable to both targeted structures. $\epsilon=1e-6$ and $\gamma=0.25$ have been assigned to compute LRP-$\epsilon$ and LRP-$\gamma$}
    \begin{tabular}{@{}l@{\hspace{0.6em}}c@{\hspace{0.4em}}c@{\hspace{0.4em}}c@{\hspace{0.4em}}c@{\hspace{0.3em}}c@{\hspace{0.3em}}c@{\hspace{0.5em}}c@{\hspace{0.5em}}c@{}}
        \\
        \toprule
        \textbf{Structure}  & \textbf{LLL} & {\textbf{MLL}} & {\textbf{HLL}} & {\textbf{FC}} & {\textbf{Softmax}} & {\textbf{\gls{mag}}} & ($\uparrow$){\textbf{$A^{PR}$}} & Top-PR($\%$)\\
        \midrule

        Attention Heads & $\epsilon$ & $\epsilon$ & $z^{+}$ & $\alpha^{2} \beta^{1}$ & $\epsilon$ & True & 0.851 & 57 \\ 
        Linear Layers & $\epsilon$ & $\alpha^{2} \beta^{1}$ & $\gamma$ & $\gamma$ & CP-LRP & True & 0.871 & 66 \\
        \midrule
        Across All &  $\epsilon$ & $\epsilon$ & $\epsilon$ & $z^{+}$ & $z^{+}$ & True & 0.851 &  \\
        \bottomrule
    \end{tabular}
\end{table*}

\begin{table*}[h!]
\label{app:tab:ablation}
    \centering
    \caption{The importance of tuning an individual parameter that is involved in constructing \gls{lrp} composite and pruning framework, can be assessed by computing the deviation in performance when that specific parameter is altered while others remain constant. The most influential ones per architecture are displayed in bold font.}
    \begin{tabular}{@{}l@{\hspace{0.6em}}c@{\hspace{0.4em}}c@{\hspace{0.4em}}c@{\hspace{0.4em}}c@{\hspace{0.3em}}c@{\hspace{0.3em}}c@{}}
        \\    
        \toprule
        \textbf{Model}  & \textbf{LLL} & {\textbf{MLL}} & {\textbf{HLL}} & {\textbf{FC}} & {\textbf{Softmax}} & {\textbf{\gls{mag}}}\\
        \midrule

        VGG-16 & 0.028 & \textbf{0.075} & 0.047 & 0.030 &   & 0.120 \\
        VGG-16-BN & 0.009 & 0.062 & \textbf{0.072} & 0.025 &   & - \\
        ResNet-18 & 0.019 & - & \textbf{0.39} & 0.17 &   & - \\
        ResNet-50 & 0.059 & \textbf{0.72} & 0.070 & 0.017 &  & 0.205 \\
        \midrule
        ViT-B-16 (Linear Layers) & \textbf{0.102} & 0.077 & 0.05 & 0.012 & 0.006 & - \\
        ViT-B-16 (Attention Heads) & 0.012 & \textbf{0.048} & 0.025 & 0.029 & 0.012 & \textbf{0.061} \\
        
        \bottomrule
    \end{tabular}
\end{table*}

\subsection{Evaluating the Composite}
\label{app:experiment:evaluate_attnlrp_lrppruner}

\begin{figure}[t]
  \centering
    \includegraphics[width=1\linewidth]{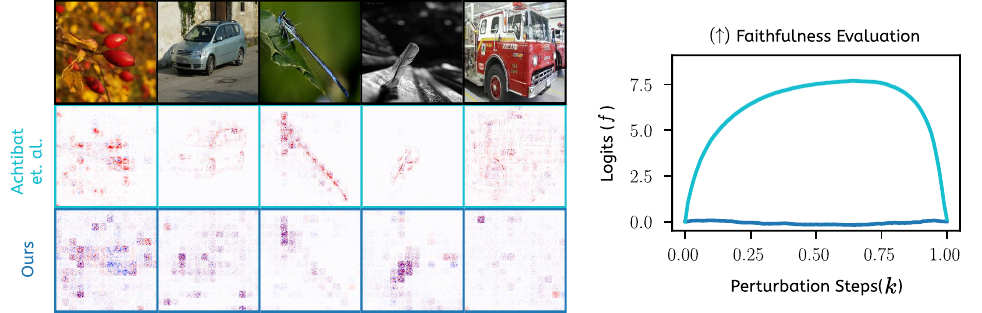}
   \caption{Qualitative experiment of ViT-B-16 on the \textit{left} shows that the composite from \cite{pmlr-v235-achtibat24a} generates more localized and less noisy explanations in the input-space rather than ours, proposed in \cref{app:tab:vit_composite}. A quantitative experiment of evaluating attribution methods on \textit{right} side of the figure validates this finding. The higher area under the curve induces a more faithful heatmap. This experiment took place by using around $1000$ samples from ImageNet and $100$ perturbation steps (see work of \cite{blücher2024decoupling} for the description of this metric).}
   \label{fig:attnlrp_lrppruner}
\end{figure}

As described in \cref{sec:methods:lrp_hyperparameters}, we proposed our pruning framework as a means of evaluating attribution methods in terms of aligning with the actual importance of latent components. However, it is noted that composites highly scored by this criterion, are not necessarily performant in generating explanation heatmaps. Our results in \cref{experiments:vit} exhibited that \gls{lrp} composite (\cref{app:sec:faithful_lrp}) suggested by \cite{pmlr-v235-achtibat24a} is not the most suitable option for pruning. However, qualitative and quantitative experiment (based on \cite{balduzzi2017shattered, bach2015pixel, pmlr-v235-achtibat24a}) in \cref{fig:attnlrp_lrppruner} demonstrates that our best composite (see \cref{app:tab:vit_composite}) is not proper to render explanation heatmap. All in all, \gls{lrp} explainers tuned to perform well in pruning, do not necessarily induce reliable heatmaps.

\section{Layer-wise Relevance Flow}
\begin{figure}[t]
  \centering
    \includegraphics[width=1\linewidth]{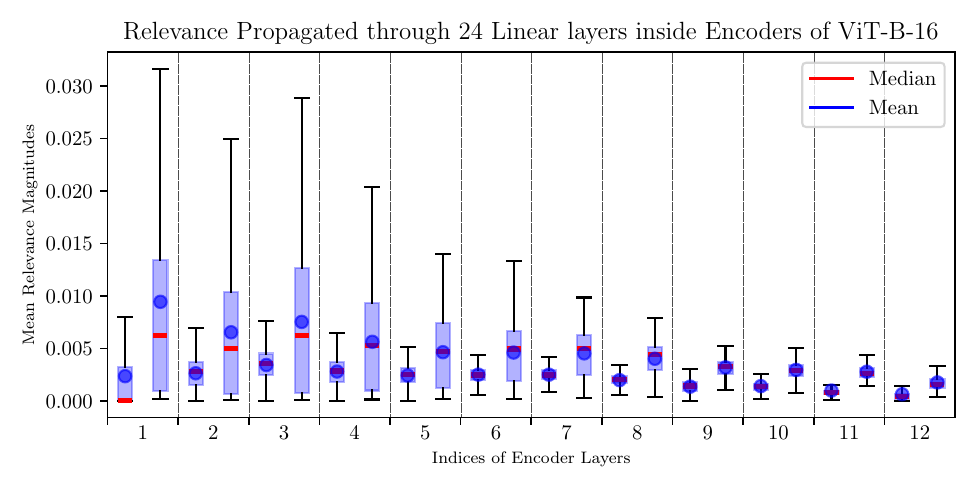}
  \caption{There are two linear layers inside 12 \texttt{Encoder} layers of ViT-B-16. Magnitude of LRP-$\epsilon$ relevance shows higher magnitudes as we approach lower layers. Interestingly, the second layers in each \texttt{Encoder} block has generally higher relevance values, which can be expected, as here 512 instead of 2048 neurons (as in the previous layer) are included. 
  }
   \label{fig:relevance_flow_vit_linear}
\end{figure}

\begin{figure}[t]
  \centering
    \includegraphics[width=1\linewidth]{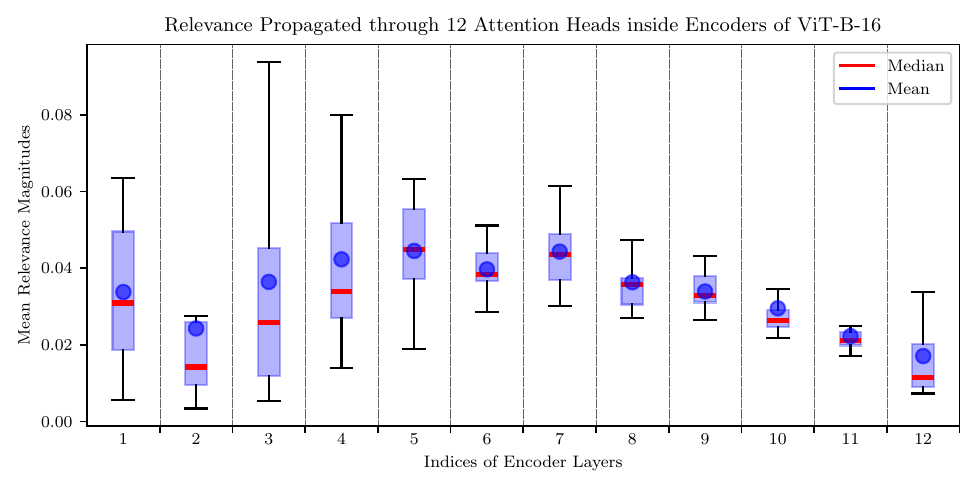}
  \caption{144 attention heads are distributed in 12 \texttt{Encoder} layers of ViT-B-16 highlight higher relevance in mid-level layers, corresponding to higher importance in decision-making.}
\label{fig:relevance_flow_vit_heads}
\end{figure}

\begin{figure}[t]
  \centering
    \includegraphics[width=1\linewidth]{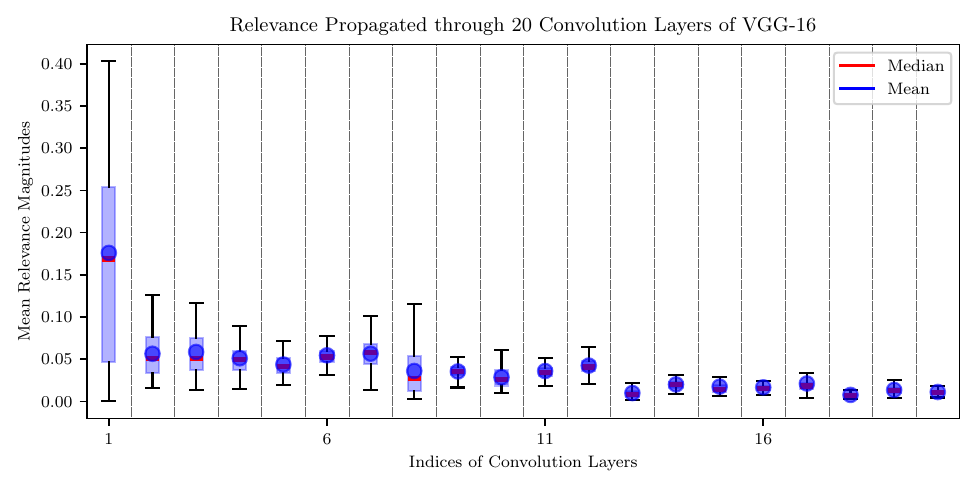}
  \caption{Generally higher relevance for mid to low level convolution filters. There are in total, 4800 convolution filters in ResNet-18 distributed in 20 layers.}
\label{fig:relevance_flow_resnet18}
\end{figure}

\begin{figure}[t]
  \centering
    \includegraphics[width=1\linewidth]{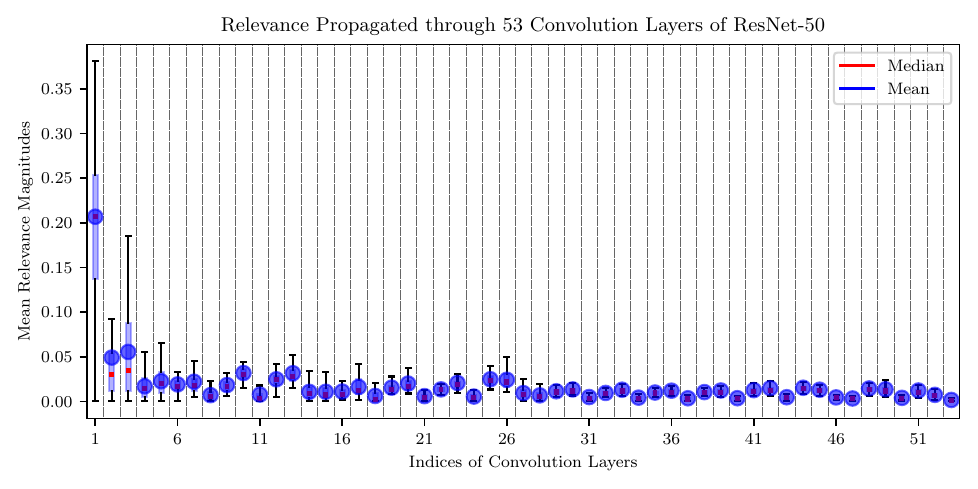}
  \caption{There are in total, 26560 convolution filters in ResNet-50 distributed in 53 layers. Higher relevance magnitude in first layers generally stems from having a few number of convolution filters there.}
   \label{fig:relevance_flow_resnet50}
\end{figure}

\begin{figure}[t]
  \centering
    \includegraphics[width=1\linewidth]{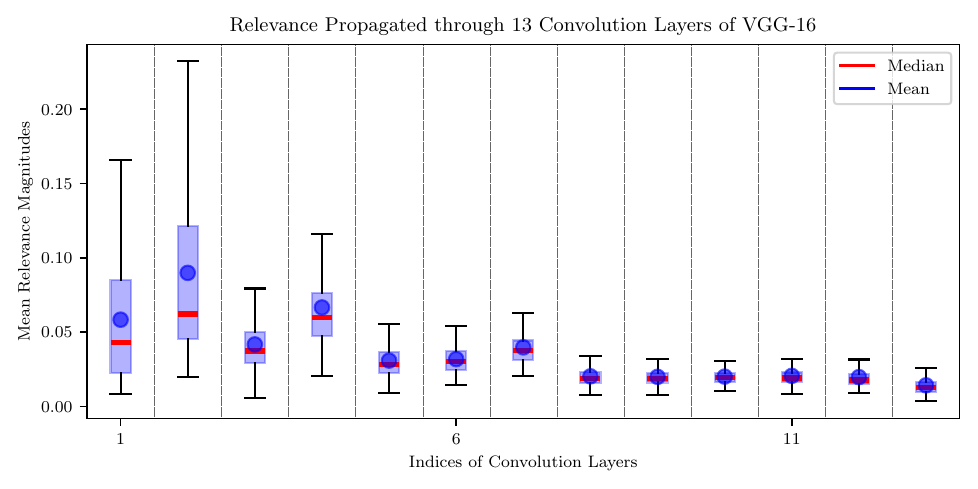}
  \caption{There are in total, 4224 convolution filters in VGG-16 distributed in 13 layers. Higher relevance magnitude is visible at mid to lower-level convolution filters.}
   \label{fig:relevance_flow_vgg16}
\end{figure}

\begin{figure}[t]
  \centering
    \includegraphics[width=1\linewidth]{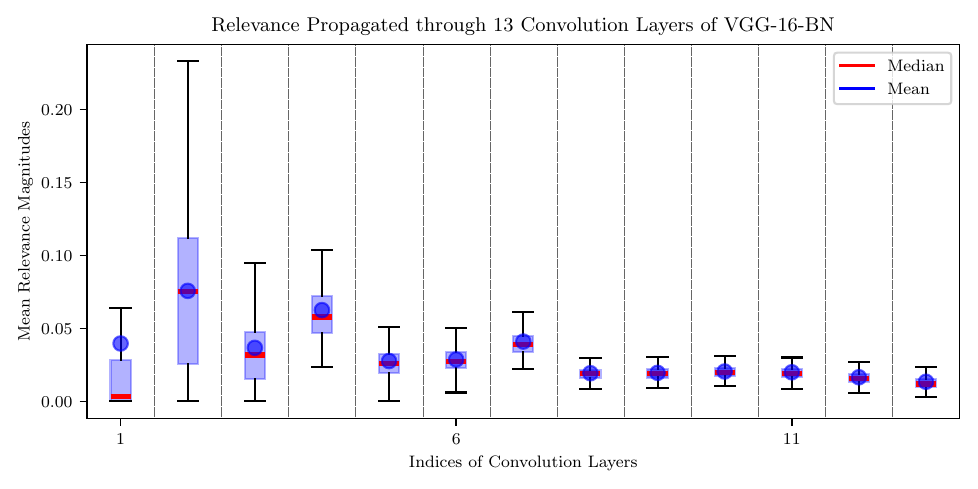}
  \caption{Similar to \cref{fig:relevance_flow_vgg16} there are in total, 4224 convolution filters in VGG-16-BN distributed in 13 layers. Higher relevance magnitude is visible at mid to lower-level convolution filters.}
\label{fig:relevance_flow_vgg16_bn}
\end{figure}

Later figures (\ref{fig:relevance_flow_vit_linear}, \ref{fig:relevance_flow_vit_heads}, \ref{fig:relevance_flow_resnet18}, \ref{fig:relevance_flow_resnet50}, \ref{fig:relevance_flow_vgg16}, \ref{fig:relevance_flow_vgg16_bn}) partially elucidate the pruning strategy by illustrating relevance magnitude of components in a layer-wise fashion.

\section{Change in Behaviour}
\begin{figure}[t]
  \centering
    \includegraphics[width=1\linewidth]{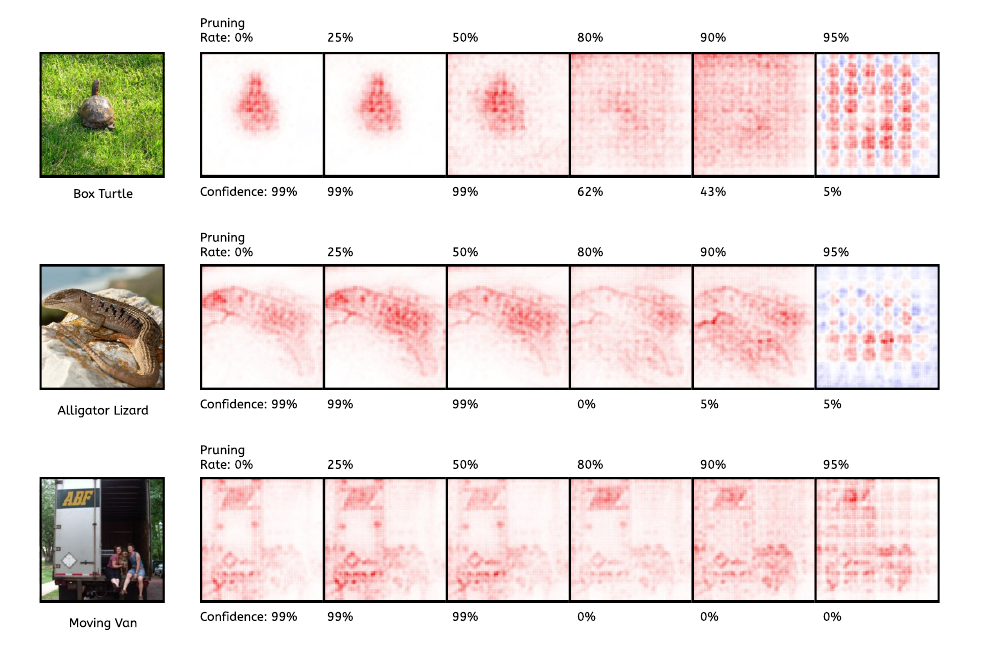}
  \caption{Changes in ResNet-18 decision-making for domain restriction of 3 randomly chosen classes (\textit{Box Turtle}, \textit{Alligator Lizard}, and \textit{Moving Van}) via LRP heatmaps (using composite described in \cref{app:sec:faithful_lrp}) indicate model failure in high compression rates are caused by high influence of surrounding objects and background despite considering the main object itself. Heatmaps of the model with high pruning rates demonstrate an effect caused by a subsampling shortcut connection becoming apparent.}
   \label{fig:heatmap_set_1}
\end{figure}

\begin{figure}[t]
  \centering
    \includegraphics[width=1\linewidth]{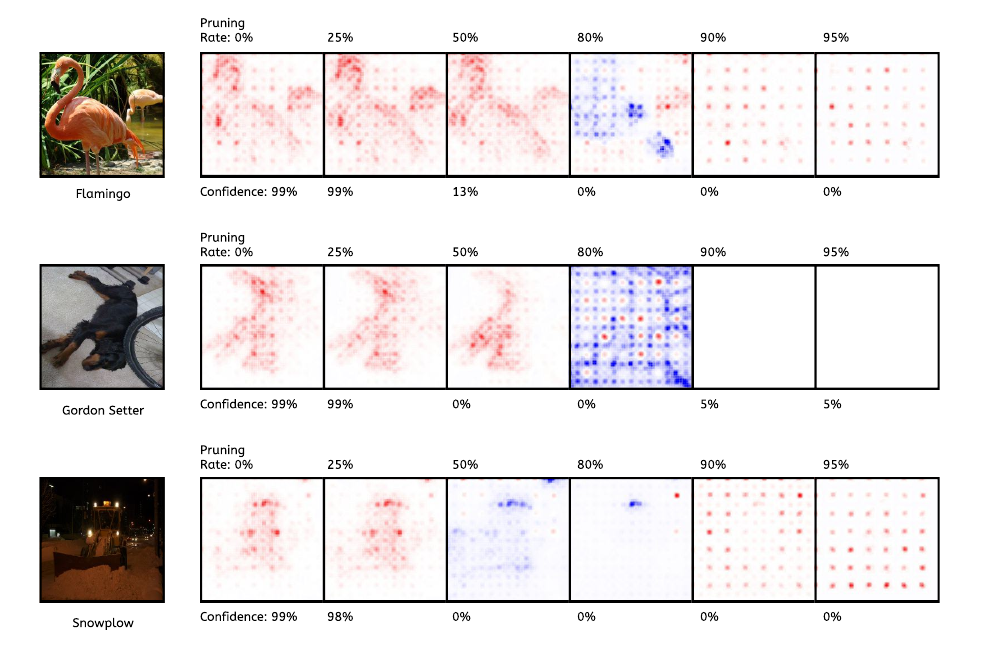}
  \caption{Changes in ResNet-50 decision-making for domain restriction of 3 randomly chosen classes (\textit{Flamingo}, \textit{Gordon Setter}, and \textit{Snowplow}) via LRP heatmaps (using composite described in \cref{app:sec:faithful_lrp}) indicate model failure in high compression rates are caused by high influence of surrounding objects and unlike ResNet-18 ignores the main object. The shortcut connections of ResNet architecture result in checkerboard pattern exhibited in the heatmaps.}
  \label{fig:heatmap_set_2}
\end{figure}

\begin{figure}[t]
  \centering
    \includegraphics[width=1\linewidth]{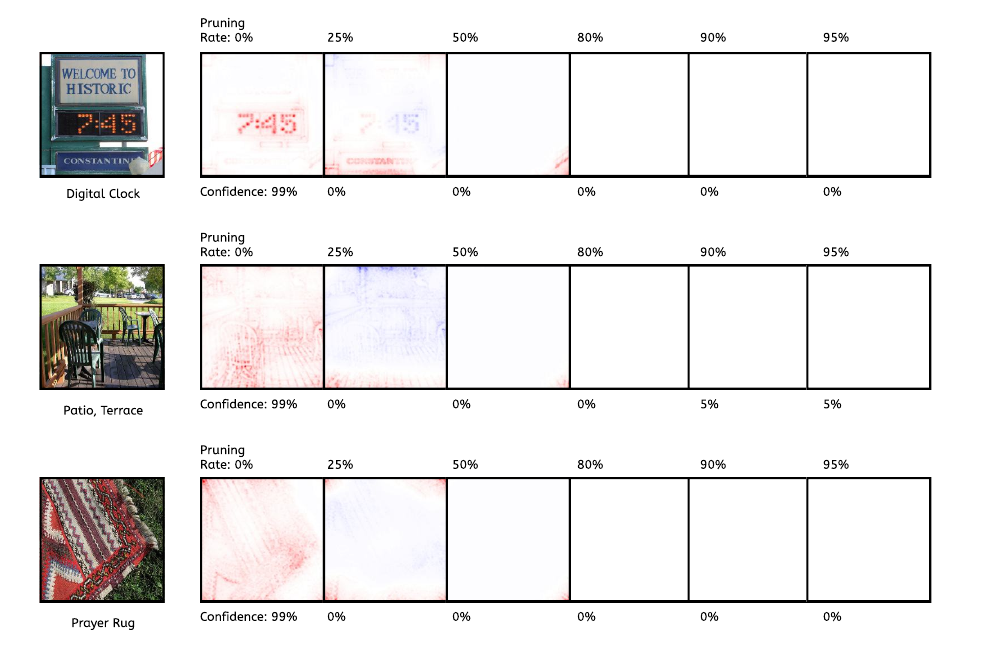}
  \caption{Changes in VGG-16-BN decision-making for domain restriction of 3 randomly chosen classes (\textit{Digital Clock}, \textit{Patio, Terrace}, and \textit{Prayer Rug}) via LRP heatmaps (using composite described in \cref{app:sec:faithful_lrp}) indicate lower pruning capabilities of this architecture compared to ResNet-18 and ResNet-50.}
  \label{fig:heatmap_set_3}
\end{figure}

In later figures (\ref{fig:heatmap_set_1}, \ref{fig:heatmap_set_2}, \ref{fig:heatmap_set_3}), we qualitatively demonstrate examples of pruning based on 3 randomly chosen classes and trace changes in the explanation heatmap of the pruned model.

\end{document}